\documentclass[runningheads]{llncs}

\usepackage{eccv}

\usepackage{eccvabbrv}

\usepackage{graphicx}
\usepackage{booktabs}

\usepackage[accsupp]{axessibility}  

\usepackage{multirow}
\usepackage{makecell}
\usepackage{adjustbox}
\usepackage{stfloats}
\usepackage{pifont}
\usepackage{subcaption}
\usepackage{tcolorbox}

\newcommand{\cmark}{\textcolor{green!70!black}{\ding{51}}}
\newcommand{\xmark}{\textcolor{red!80!black}{\ding{55}}}

\newcommand{\posd}[1]{\textbf{\textcolor{green!70!black}{#1}}}
\newcommand{\negd}[1]{\textbf{\textcolor{red!80!black}{#1}}}

\newcommand{\first}[1]{\textbf{#1}}
\newcommand{\second}[1]{\underline{#1}}

\usepackage{hyperref}

\usepackage{orcidlink}

\begin{document}

\title{PanoGrounder: Bridging 2D and 3D with Panoramic Scene Representations \\ for VLM-based 3D Visual Grounding} 

\titlerunning{PanoGrounder}

\author{Seongmin Jung\inst{1}$^{\star}$\orcidlink{0009-0002-5281-4375} \and
Seongho Choi\inst{1,2}$^{\star}$\orcidlink{0009-0007-4458-6823} \and
Gunwoo Jeon\inst{1}\orcidlink{0009-0008-1410-9556} \and
Minsu Cho\inst{3}\orcidlink{0000-0001-7030-1958} \and \\
Jongwoo Lim\inst{1}\orcidlink{0000-0002-2814-4765}}

\authorrunning{S.~Jung et al.}

\institute{Seoul National University, South Korea\\
\email{\{sm18570, sh.choi, gunw0, jongwoo.lim\}@snu.ac.kr}\and
Robotics Lab, Hyundai Motor Company, South Korea\and
Pohang University of Science and Technology (POSTECH), South Korea\\
\email{mscho@postech.ac.kr}\\
\href{https://choiseongho-h.github.io/PanoGrounder}{https://choiseongho-h.github.io/PanoGrounder}}

\maketitle
\let\thefootnote\relax\footnotetext{$^{\star}$ Equal contribution.}

\begin{abstract}
3D Visual Grounding (3DVG) is a critical bridge from vision-language perception to robotics, requiring both language understanding and 3D scene reasoning. Traditional supervised models leverage explicit 3D geometry but exhibit limited generalization, owing to the scarcity of 3D vision-language datasets and the limited reasoning capabilities compared to modern vision-language models (VLMs). We propose a generalizable 3DVG framework, \textbf{PanoGrounder}, that couples multi-modal panoramic representation with pretrained 2D VLMs for strong vision-language reasoning. Panoramic renderings, augmented with 3D semantic and geometric features, serve as an intermediate representation between 2D and 3D, and offer two major benefits: (i) they can be directly fed to VLMs with minimal adaptation and (ii) they retain long-range object-to-object relations thanks to their 360-degree field of view. We devise a three-stage pipeline that places a compact set of panoramic viewpoints considering the scene layout and geometry, grounds a text query on each panoramic rendering with a VLM, and fuses per-view predictions into a single 3D bounding box via lifting. Our approach achieves state-of-the-art results on ScanRefer and Nr3D, and demonstrates strong generalization to unseen 3D datasets and text rephrasings.
\end{abstract}

\section{Introduction}

\begin{figure}[t]
    \centering
    \includegraphics[width=\columnwidth]{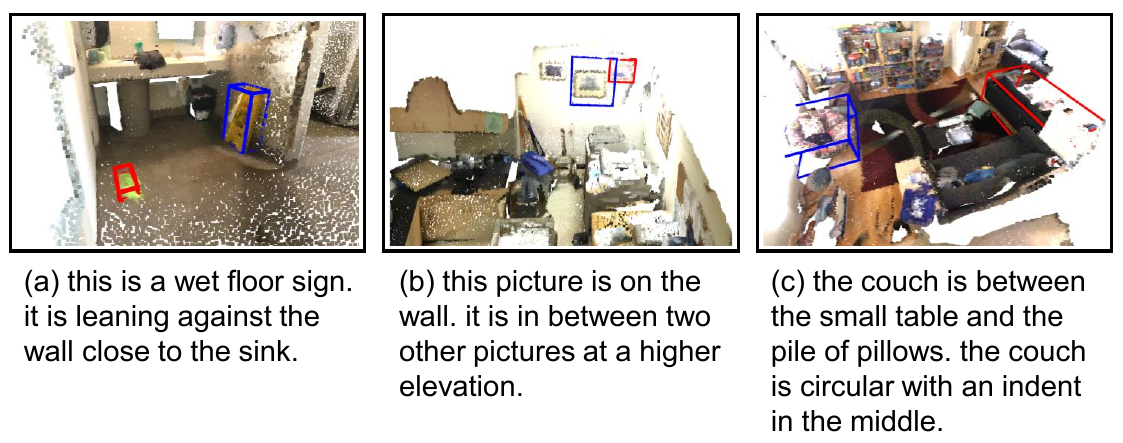}
    \caption{\textbf{PanoGrounder} localizes referred objects in 3D scenes by leveraging panoramic renderings as a 2D--3D interface. Compared to 3D-VisTA~\cite{zhu20233d_vista} (red), PanoGrounder (blue) accurately grounds (a) rare objects, (b) small objects with fine-grained spatial relations, and (c) complex queries involving multiple inter-object relationships.}
    \label{fig:hook}
\end{figure}

3D Visual Grounding (3DVG) aims to localize a target object in a 3D scene from a free-form natural language query, sitting at the interface of natural language understanding and 3D scene understanding~\cite{chen2020scanrefer, achlioptas2020referit3d}. As a core capability for embodied AI, 3DVG underpins applications in augmented reality, vision-language navigation, and robotic perception and manipulation~\cite{anderson2018vln}. The task requires \emph{language understanding}, which extracts attributes and relations from the instruction, and \emph{3D scene understanding}, which grounds those cues in metric geometry to find the target object~\cite{liu20253dvgsurvey}.

Traditional 3DVG systems typically employ separate encoders for text and point clouds, followed by cross-modal fusion~\cite{liu20253dvgsurvey}. By operating directly on the 3D scene, these models have achieved strong performance on standard benchmarks~\cite{chen2020scanrefer, achlioptas2020referit3d}. Nevertheless, this design has limitations in both \textit{language comprehension} and \textit{3D vision generalization}. On the language side, most works rely on BERT~\cite{devlin2019bert} or CLIP~\cite{radford2021clip}-style encoders, which offer more limited compositional and spatial reasoning capabilities than modern VLMs~\cite{niven2019bertstat, yuksekgonul2023clipbow, zhang2023bertreason, lewis2024clipcompo, wang2024cogvlm, liu2023llava}, making paraphrases and relation-heavy descriptions difficult to handle. On the 3D vision side, these models are trained from scratch on limited-scale 3DVG datasets~\cite{chen2020scanrefer, achlioptas2020referit3d}, which restricts their generalization to scenes, categories, and linguistic expressions outside of the training distribution.
To leverage stronger linguistic reasoning, recent work introduces VLMs into the 3DVG pipeline. However, transferring 2D capacity to 3D remains non-trivial. Recent VLM-based methods~\cite{li2025seeground, xu2024vlmgrounder} use perspective (pinhole) images as a 2D--3D interface, but the limited field of view fails to capture a holistic spatial context. Moreover, reliance on large proprietary models makes these approaches computationally prohibitive and difficult to fine-tune. This calls for a simpler, more efficient intermediate representation that connects 2D VLMs with 3D reasoning.

In this paper, we introduce \textbf{PanoGrounder}, which uses panoramic renderings as an explicit intermediate representation between 2D and 3D modalities. Unlike perspective images, panoramas cover a $360^\circ$ field of view, capturing holistic spatial context within a single image while remaining fully compatible with VLMs. By operating on renderable representations rather than raw point clouds, this design also relaxes the high-quality 3D input requirement common in traditional 3DVG pipelines---requiring only casually captured RGB frames and an off-the-shelf SfM/NVS pipeline.
PanoGrounder operates in three stages: (i) a compact set of panoramic viewpoints is selected and multi-modal panoramas---RGB, semantic, and range---are rendered for richer contextual cues; (ii) a pretrained VLM processes each panorama to predict a 2D bounding box in pixel coordinates; and (iii) the 2D grounding outputs are lifted into metric 3D space and fused across views. To incorporate semantic and geometric context from the multi-modal renderings, the VLM is augmented with a lightweight adapter, fine-tuned in the panoramic domain.

As shown in \cref{fig:hook}, PanoGrounder achieves accurate grounding on challenging cases involving rare objects, fine-grained spatial relations, and complex multi-object queries. Quantitatively, it achieves state-of-the-art performance on ScanRefer~\cite{chen2020scanrefer} and Nr3D~\cite{achlioptas2020referit3d}. We further assess (i) \textit{scene generalization} on ARKitScenes+SceneVerse~\cite{dehghan2021arkitscenes, jia2024sceneverse} and 3RScan+RIORefer~\cite{wald2019rio, miyanishi2024cross3dvg} and (ii) \textit{text generalization} with a modified version of ScanRefer queries. Across all settings, our method exhibits strong generalization compared to fully supervised baselines.

Our main contributions are as follows:
\begin{itemize}
  \item We introduce a novel 3DVG approach that treats panoramic renderings as a 2D--3D interface for pretrained 2D VLMs. This design preserves scene-wide context while enabling powerful vision-language reasoning with only minimal modifications around the VLM.
  
  \item We propose an effective 3DVG model, PanoGrounder, that injects \emph{semantic} and \emph{geometric} features into the VLM's vision encoder via a multi-modal feature adapter and introduces an Earth Mover's Distance (EMD) loss that provides distance-aware supervision for more accurate localization.
  
  \item Our method, PanoGrounder, achieves state-of-the-art performance on ScanRefer and Nr3D. It further demonstrates strong generalization to unseen scenes and diverse text rephrasings.
\end{itemize}
\section{Related Work}

\subsection{3D Visual Grounding}

\subsubsection{3D-based 3DVG.}

Traditional 3D-based methods are fully supervised, using separate encoders for text and 3D point clouds and fusing them via a cross-modal module. Two-stage approaches~\cite{achlioptas2020referit3d, chen2022vil3drel, zhu20233d_vista, zhu2024pq3d, chang2024mikasa, huang2025viewsrd} follow a proposal-and-selection paradigm, where a 3D detector first proposes candidate objects and the model selects the best match. ViewSRD~\cite{huang2025viewsrd} further improves language understanding by decomposing complex queries into simpler clauses via an LLM. One-stage methods~\cite{jain2022butddetr, wu2023eda, qian2024mcln, jain2025univlg} directly regress 3D bounding boxes or masks conditioned on the query; BUTD-DETR~\cite{jain2022butddetr} introduces language and objectness guidance into a DETR-style decoder, while MCLN~\cite{qian2024mcln} extends it with two parallel decoders for box-level and mask-level prediction, enhancing overall localization consistency.
Despite these advances, most methods rely on BERT or CLIP text encoders and are trained from scratch on limited-scale 3DVG datasets, constraining both spatial reasoning and generalization to unseen scenes and expressions. Furthermore, their reliance on high-quality, human-cleaned point clouds limits practical applicability, as performance degrades significantly when using raw RGBD-projected point clouds~\cite{jain2025univlg}. In contrast, PanoGrounder leverages pretrained VLMs via a panoramic 2D--3D interface, enabling stronger language understanding and more robust generalization without requiring curated 3D inputs.

\subsubsection{2D-based 3DVG.}

Without relying on a global 3D scene point cloud, a line of works performs 3DVG directly from one or a small set of RGB(-D) views. Refer-it-in-RGBD~\cite{liu2021referitinrgbd} operates on single-view RGB-D and employs a coarse-to-fine framework to recover the full 3D extent of partially observed targets. Mono3DVG~\cite{zhan2024mono3dvg} tackles single-view monocular RGB in autonomous-driving scenes, leveraging a lightweight depth predictor. Recent zero-shot approaches~\cite{yang2024llmgrounder, yuan2024zsvg3d, xu2024vlmgrounder, li2025seeground} prompt an LLM/VLM with multi-turn instructions combining images and text to inject scene context. However, all these methods share a common limitation: the choice of viewing direction critically affects what scene content is captured, making it difficult to capture holistic spatial context across the scene. Zero-shot variants additionally rely on large models not designed for task-specific fine-tuning, making adaptation costly. In contrast, PanoGrounder uses panoramic renderings to capture scene-wide context within a single image, and its adapter-based design enables efficient end-to-end fine-tuning on task-specific data.

\subsection{Multi-Modal 3D Perception}

A complementary line of work jointly trains models on multiple tasks, such as 3D visual grounding, 3D scene captioning, and Visual Question Answering (VQA), to align 3D and language spaces. Several methods~\cite{zhu20233d_vista, li2023uni3dl, zhu2024pq3d} adopt transformer-based alignment where task-specific heads branch from a shared backbone. UniVLG~\cite{jain2025univlg} further leverages abundant 2D data via a neural 2D-to-3D lifting model to transfer supervision into 3D. Beyond supervised alignment, 3D-R1~\cite{huang20253dr1} explores RLHF-style policy optimization to refine instruction following in 3D contexts. Recently, Multi-modal Large Language Models (MLLMs) have been investigated by feeding 3D-aware tokens into pretrained VLM backbones~\cite{liu2023llava}. Scene-LLM~\cite{fu2025scenellm} forms hybrid voxel-point tokens from multi-view features~\cite{radford2021clip} and linearly projects them into the LLM space. Chat-Scene~\cite{huang2024chatscene} uses object proposals with per-instance Object Identifier tokens to fuse 2D/3D object features for unified object-token reasoning. LLaVA-3D~\cite{zhu2025llava3d} augments 2D patch tokens with 3D position embeddings to create 3D-aware patches for the LLM.
These approaches require task-specific 3D token designs and substantial architectural modifications to bridge 3D and language spaces. In contrast, PanoGrounder bridges 3D and language spaces through panoramic renderings, enabling direct reuse of pretrained 2D VLMs with minimal architectural overhead.
\begin{figure*}[t]
    \centering
    \includegraphics[width=\textwidth]{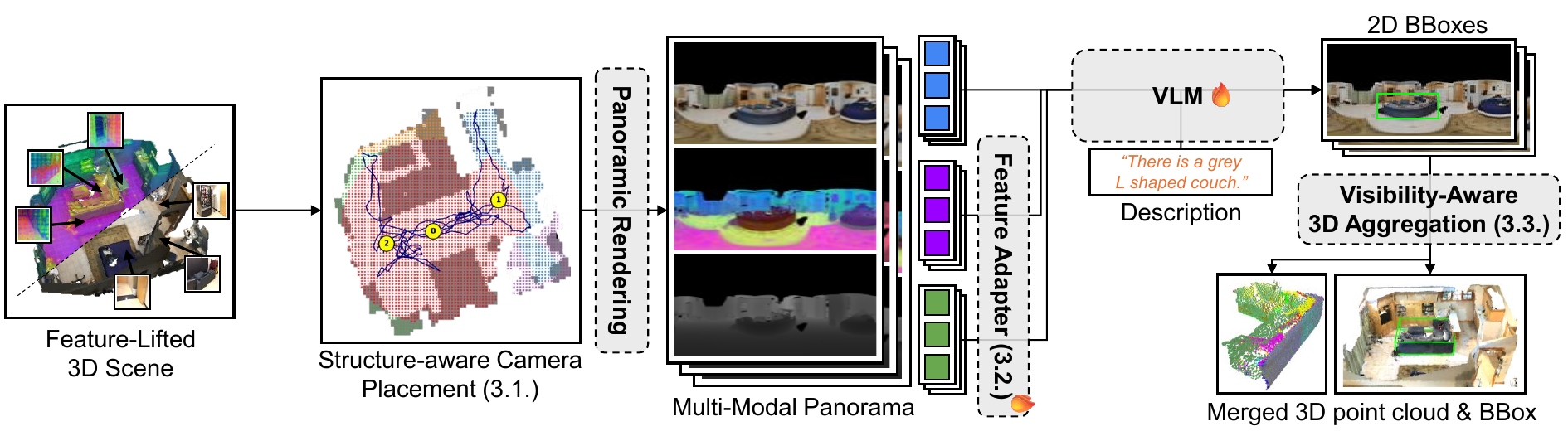}
    \caption{\textbf{Overall pipeline of PanoGrounder.} We first select a compact set of informative panoramic viewpoints via Structure-aware Camera Placement and render multi-modal panoramas (RGB, lifted semantic features, and range). Each view is processed by an off-the-shelf VLM augmented with our multi-modal feature adapter to produce 2D bounding boxes from the text query. Finally, per-view 2D predictions are lifted and fused with visibility-aware 3D aggregation to yield a single 3D bounding box for the referred object.}
    \label{fig:pipeline}
\end{figure*}

\section{Method}
\label{sec:method}

We assume the 3D scene is given as a renderable representation (e.g., a triangle mesh or 3D Gaussian Splatting~\cite{kerbl20233dgs}). Given such a representation and a text query, our goal is to predict a 3D bounding box of the referred object. Our key idea is to use panoramic renderings as an intermediate representation that bridges 2D vision-language understanding and 3D spatial reasoning.

As illustrated in \cref{fig:pipeline}, we first select a compact set of informative viewpoints (\cref{sec:scp}) and render multi-modal panoramas---RGB, geometric, and semantic feature maps---at each location (\cref{sec:mmpr}). These are fed into a VLM augmented with lightweight adapters that inject geometric and semantic cues (\cref{sec:3dca}), producing per-view 2D bounding box predictions. The predictions are then lifted and fused into a 3D bounding box via visibility-aware 3D aggregation (\cref{sec:va3a}). We train the model with cross-entropy combined with an Earth Mover's Distance loss for distance-aware supervision (\cref{sec:trobj}).

\subsection{Structure-Aware Camera Placement}
\label{sec:scp}

Panoramic cameras capture an omnidirectional view of the scene, bypassing the need to predict a specific viewing orientation, so we only select \emph{locations}. We start by estimating the floor via RANSAC~\cite{fischler1981ransac} and place a regular grid $\mathcal{G}{=}\{p_i\}$ with $h{=}10\,\mathrm{cm}$ spacing across the scene's 2D floor footprint. Each grid point $p_i$ serves as a candidate panoramic camera location, with its height set to the average height of the raw cameras used during scene reconstruction.

\subsubsection{Scoring factors.}
To score each candidate $p \in \mathcal{G}$, we construct three factors:
\begin{itemize}
\item \textbf{Ray coverage} $A(p)\in[0,1]$: fraction of other grid points visible from $p$ within $r_{\max}{=}3\,\mathrm{m}$, counted via obstacle-free projection onto the panoramic image plane.
\item \textbf{Distance-to-surface} $D_{\mathrm{surf}}(p)$: Euclidean distance from $p$ to the nearest scene geometry (walls or furniture).
\item \textbf{Distance-to-trajectory} $D_{\mathrm{traj}}(p)$: Euclidean distance from $p$ to the nearest raw RGB camera center. This term is optional---when no prior camera trajectory is available, it is simply omitted.
\end{itemize}

\subsubsection{Score and selection.}
We rank candidates using
\begin{equation}
S(p) = \frac{A(p)\, D_{\mathrm{surf}}(p)}{D_{\mathrm{traj}}(p) + \varepsilon},
\quad \varepsilon{=}10^{-3},
\end{equation}
which balances area coverage, obstacle clearance, and trajectory proximity. Starting from the highest-scoring point, we greedily select cameras. Once a camera is selected, all grid points visible from it are marked as `covered', effectively setting their contribution to the ray coverage $A(p)$ to zero for all remaining candidates. This sequential selection is repeated until at least $90\%$ of the scene's grid points are covered.

\subsection{Multi-Modal Panoramic VLM}
\label{sec:mmpr}

\begin{figure}[t]
    \centering
    \includegraphics[width=0.8\columnwidth]{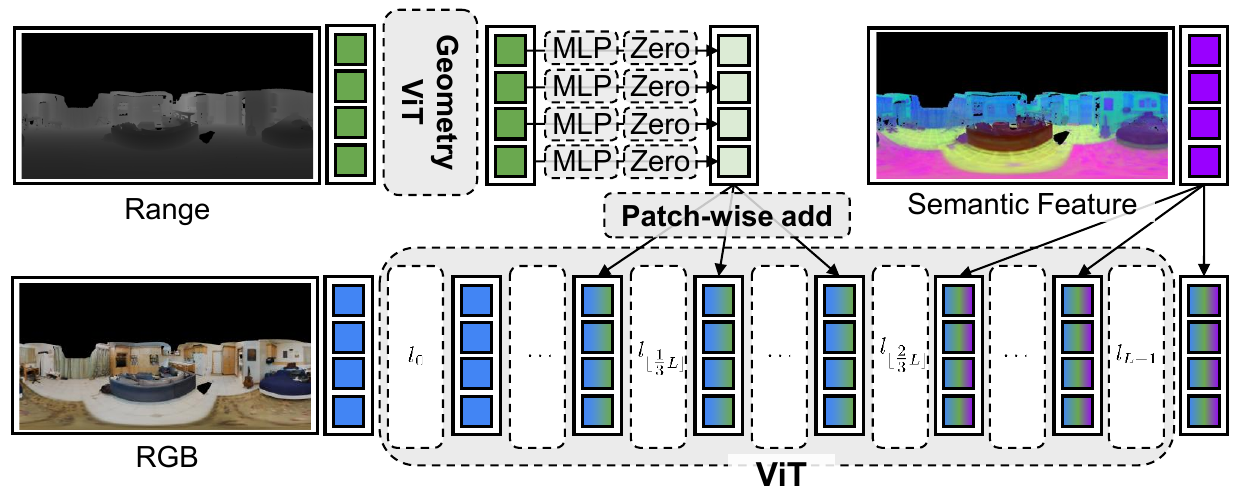}
    \caption{\textbf{Multi-modal feature adapter.} We inject panoramic \emph{geometric} (range) and \emph{semantic} (multi-view) features into selected ViT layers of the VLM. Each modality is processed by a lightweight adapter (2-layer MLP followed by a $1{\times}1$ convolution), and the resulting output is added patch-wise to the ViT tokens. The $1{\times}1$ convolution weights and bias are initialized to zero.}
    \label{fig:adapter}
\end{figure}

For each selected viewpoint, indexed by $k$, we render a panoramic RGB image $I_k \in \mathbb{R}^{H \times W \times 3}$ using equirectangular projection. To better handle occlusions in cluttered indoor scenes and to provide explicit 3D information for spatial reasoning, we additionally extract geometric and semantic feature maps from the same panoramic cameras and inject them into the VLM alongside RGB via lightweight adapters (\cref{fig:adapter}). All modalities share a common token grid of $N_f = H_f \times W_f$ patches ($H_f$ and $W_f$ are the patch counts along the image height and width), so that the injected features align patch-wise with the VLM's RGB tokens.

\subsubsection{Geometric feature map.} For each selected viewpoint, we render a range (depth) map $D_k \in \mathbb{R}^{H \times W}$. Inspired by \cite{cheng2024spatialrgpt}, $D_k$ is fed to a
pretrained ViT~\cite{oquab2023dinov2} to obtain a $d_{\mathrm{geo}}$-dimensional geometric feature map
$\mathbf{f}_{\mathrm{geo}}\in\mathbb{R}^{N_f \times d_{\mathrm{geo}}}$ aligned with this token grid.

\subsubsection{Multi-view fused semantic feature map.}
We extract dense semantic patch features from a frozen ViT encoder~\cite{oquab2023dinov2} applied to the original RGB views of the dataset. Using known camera intrinsics and extrinsics, each mesh vertex visible in a view is assigned the $d_{\mathrm{sem}}$-dimensional feature of the patch it projects into. Since a vertex is typically visible from multiple views, the features it receives are averaged and re-rendered onto the selected panoramic cameras to produce a semantic feature map $\mathbf{f}_{\mathrm{sem}}\in\mathbb{R}^{N_f \times d_{\mathrm{sem}}}$.

\subsubsection{Feature adapter.}
\label{sec:3dca}
We inject both geometric and semantic features into the VLM's vision encoder via a shared adapter design: a 2-layer MLP followed by a $1{\times}1$ convolution whose weights and bias are initialized to zero, following Zero-Convolution~\cite{zhang2023controlnet}. At initialization the adapter output is exactly zero, so the VLM behaves identically to its pretrained form; fine-tuning then lets the adapters encode task-relevant signals without distorting the original representation space. Geometric features are injected into \emph{mid-level} layers ($l \in [L/3,\,2L/3)$) to provide a spatial scaffold, whereas semantic features are injected into \emph{later} layers ($l \in [2L/3,\,L]$) to supply high-level contextual cues:
\begin{equation}
\mathbf{X}_{l,n} \leftarrow \mathbf{X}_{l,n} + \text{ZeroConv}_{m}\!\left(\text{MLP}_{m}(\mathbf{f}_{m,n})\right),
\end{equation}
where $m\in\{\mathrm{geo},\mathrm{sem}\}$ indexes the modality, $L$ is the number of VLM encoder layers, $l$ and $n$ are the layer and token indices, and $\mathbf{f}_{m,n}$ is the $n$-th token of $\mathbf{f}_m$.

\subsubsection{Training supervision.}
To obtain ground-truth 2D bounding boxes, we render a per-pixel instance-ID map at each selected camera via the same equirectangular projection used for RGB. Each 3DVG dataset provides (text query, target object ID) pairs; we compute the tightest 2D bounding box enclosing all pixels of the target instance, yielding \emph{image--text--box} triplets without any manual labeling.

\subsection{Visibility-Aware 3D Aggregation}
\label{sec:va3a}

Let $b_k$ denote the 2D bounding box predicted by the VLM for the $k$-th viewpoint. This stage leverages cross-view consistency to filter out erroneous individual 2D predictions and accurately localize the target object in 3D.

\subsubsection{Mask-to-3D lifting.}
Given $I_k$ and $b_k$, we employ an off-the-shelf segmentation model (e.g., \textsc{SAM}~\cite{kirillov2023sam}) to extract a precise object mask $M_k$. The masked pixels are then unprojected into world coordinates using $D_k$ and the camera extrinsics of view $k$, yielding a per-view 3D point set $\mathcal{P}_k$.

\subsubsection{Anchor view selection.}
To identify the most reliable viewpoint, we project each candidate point set $\mathcal{P}_k$ onto all other views $t \neq k$ and compute the tight 2D bounding box $\hat{b}_{k \to t}$ enclosing the projected points.  We then formulate a cross-view consistency score:
\begin{equation}
\mathrm{score}(k) = \sum_{t \in \mathcal{K} \setminus \{k\}} \mathrm{IoU}\!\left(\hat{b}_{k \to t}, b_t\right),
\end{equation}
where $\mathcal{K}$ is the set of all selected viewpoints, and the \emph{anchor view} is the one maximizing this score, $k^* = \arg\max_k \mathrm{score}(k)$. This formulation explicitly favors predictions that exhibit strong geometric consensus across multiple perspectives.

\subsubsection{Multi-view fusion.}
We aggregate the individual point sets into a global cloud $\mathcal{P} = \bigcup_{k \in \mathcal{K}} \mathcal{P}_k$ and apply statistical outlier removal to mitigate noise. The denoised points are then re-projected onto the anchor view $k^*$. Points whose projections fall outside the anchor bounding box $b_{k^*}$ are discarded, resulting in a visibility-filtered set $\mathcal{P}_{\mathrm{vis}}$. Finally, we fit an axis-aligned bounding box to $\mathcal{P}_{\mathrm{vis}}$ to yield the final 3D localization.

For evaluations requiring pre-defined candidate boxes (e.g., ReferIt3D), we introduce a \emph{two-stage} variant of our framework to ensure fair comparison. This variant preserves the core algorithmic modules, with comprehensive details provided in the supplementary material.

\subsection{Training Objective}
\label{sec:trobj}

Following CogVLM~\cite{wang2024cogvlm}, each bounding box coordinate is normalized to $[000,999]$, discretized into 3 digits, and predicted as a sequence of digit tokens $\{0,\dots,9\}$. We supervise the autoregressive decoder with token-level cross-entropy under teacher forcing. However, standard cross-entropy treats each digit as an independent category, ignoring numerical proximity. Inspired by~\cite{ren2023emo}, we add an auxiliary EMD (1-Wasserstein distance) loss to inject awareness of the underlying numerical ordering.

For the $i$-th digit, let $X_i \in \{0,\dots,9\}$ be the ground-truth digit and $P^{(i)}(x)$ the predicted probability for digit $x$. The per-digit CE and EMD losses are:
\begin{align}
\mathcal{L}_{\text{CE}}^{(i)} &= -\log P^{(i)}(X_i), &
\mathcal{L}_{\text{EMD}}^{(i)} &= \omega_i \sum_{x=0}^{9} P^{(i)}(x)\, |X_i - x|,
\end{align}
where $\omega_i$ is a place-value weight (e.g., $10^2,10^1,10^0$ for hundreds/tens/ones). The EMD term penalizes larger deviations more heavily, encouraging probability mass to concentrate near the correct digit. The final objective combines both over all 3 digits:
\begin{equation}
\mathcal{L}_{\text{total}} = \sum_{i=1}^{3} \left( \mathcal{L}_{\text{CE}}^{(i)} + \lambda\, \mathcal{L}_{\text{EMD}}^{(i)} \right),
\end{equation}
where $\lambda$ balances the two terms.

To better encourage the model to leverage the injected geometric features, we augment the training data with auxiliary geometric QA pairs, inspired by \cite{cheng2024spatialrgpt}. Our QA samples are generated on-the-fly by randomly sampling two pixels on a panorama and asking the model to identify which has a larger 3D coordinate along a given axis. These auxiliary samples are supervised with cross-entropy only (details in the supplementary material).
\section{Experiments}

\subsection{Experimental Setup}
\label{sec:expsetup}

\subsubsection{Datasets.}

We evaluate on two widely used 3D visual grounding benchmarks: ScanRefer~\cite{chen2020scanrefer} and ReferIt3D (Nr3D, Sr3D)~\cite{achlioptas2020referit3d}. ScanRefer annotates 51,583 human-written referring expressions across 800 ScanNet~\cite{dai2017scannet} scenes. Nr3D contains 41,503 human utterances over 707 ScanNet scenes spanning 76 object classes, while Sr3D consists of 83,572 template-based spatial descriptions, also referring to 76 object classes.

To assess cross-dataset generalization on unseen settings, we use ARKitScenes~\cite{dehghan2021arkitscenes} scenes paired with SceneVerse~\cite{jia2024sceneverse} human-written referring expressions only (we exclude automatically generated texts), and 3RScan~\cite{wald2019rio} scenes paired with RIORefer~\cite{miyanishi2024cross3dvg} human-written referring expressions. ARKitScenes provides 5,047 captures over 1,661 unique indoor scenes, and SceneVerse includes ARKitScenes among its sources with both human and generated texts. 3RScan comprises 1,482 scans of 478 indoor environments with instance-level annotations, and RIORefer contributes 63,602 human descriptions for 1,380 3RScan scans.

\subsubsection{Evaluation Metrics.}

On ScanRefer, we report Acc@0.25 overall and on the \emph{Unique}/\emph{Multiple} subsets~\cite{chen2020scanrefer}. \emph{Unique} denotes queries with no same-class distractor, while \emph{Multiple} denotes the presence of same-class distractors. On ReferIt3D (Nr3D, Sr3D), we report Top-1 accuracy overall and on four subsets: \emph{Easy}/\emph{Hard} (exactly one vs.\ two or more same-class distractors) and \emph{View-Dependent}/\emph{View-Independent}~\cite{achlioptas2020referit3d}, indicating whether resolving the description requires a specific viewpoint (e.g., ``left/right''). For the generalization experiments (\cref{sec:generalization}), we report Top-1 accuracy given ground-truth object instances and report \emph{Unique} and \emph{Multiple} splits using the same criterion as ScanRefer.

\subsubsection{Implementation Details.}

We adopt CogVLM-17B~\cite{wang2024cogvlm} as our 2D grounding backbone, pretrained on large-scale image-text data. We fine-tune the model with LoRA, optimized by Adam with a batch size of 64 and a cosine-decay learning rate starting at $1\times10^{-4}$; all remaining hyperparameters follow the official CogVLM implementation. We train for 5 epochs counted scene-centrically (all panoramic views of a scene form a single pass); since each referring expression is seen from $\approx2.4$ viewpoints on average, this corresponds to roughly 12 text-centric epochs. For data augmentation, we apply random in-place camera yaw rotations, implemented as horizontal circular shifts (wrap-around) of the panorama. For test time augmentation, we generate four augmented views by rotating the camera in-place by $90^\circ$ increments and perform independent inference on each. For a fair comparison with prior work, all results in the main paper are obtained from mesh-rendered panoramas. However, to better reflect practical use cases, results utilizing 3DGS-rendered panoramas are provided in the supplementary material.

\subsection{3D Visual Grounding Results}

\begin{table*}[t]
\centering
\caption{\textbf{Evaluation on Nr3D, Sr3D~\cite{achlioptas2020referit3d}, and ScanRefer~\cite{chen2020scanrefer}.} To ensure a fair comparison, we group methods by training regime: models trained on a single benchmark versus those trained on mixed datasets (e.g., ScanRefer + ReferIt3D). \textit{S+R} denotes our model trained jointly on ScanRefer and ReferIt3D (Nr3D/Sr3D). For ScanRefer, we report Accuracy@0.25 (IoU). $^\dagger$Results trained on that single dataset only.}
\label{tab:benchmark}
\resizebox{\textwidth}{!}{
\begin{tabular}{llcccccccccccccc}
\toprule
& \multirow{2.5}{*}{\textbf{Method}} & \multicolumn{5}{c}{\textbf{Nr3D}} & \multicolumn{5}{c}{\textbf{Sr3D}} & \multicolumn{3}{c}{\textbf{ScanRefer}} \\
\cmidrule(lr){3-7} \cmidrule(lr){8-12} \cmidrule(lr){13-15}
& & \textbf{Easy} & \textbf{Hard} & \textbf{VD} & \textbf{VID} & \textbf{Overall} & \textbf{Easy} & \textbf{Hard} & \textbf{VD} & \textbf{VID} & \textbf{Overall} & \textbf{Unique} & \textbf{Multiple} & \textbf{Overall} \\
\midrule
\multirow{12}{*}{\rotatebox{90}{\textbf{Single Dataset}}}
& BUTD-DETR~\cite{jain2022butddetr} & 60.7 & 48.4 & 46.0 & 58.0 & 54.6 & 68.6 & 63.2 & 53.0 & 67.6 & 67.0 & 84.2 & 46.6 & 52.2 \\
& ViL3DRel~\cite{chen2022vil3drel} & 70.2 & 57.4 & 62.0 & 64.5 & 64.4 & 74.9 & 67.9 & 63.8 & 73.2 & 72.8 & 81.6 & 40.3 & 47.9 \\
& 3D-VisTA$^\dagger$~\cite{zhu20233d_vista} & 65.9 & 49.4 & 53.7 & 59.4 & 57.5 & 72.1 & 63.6 & 57.9 & 70.1 & 69.6 & 77.4 & 38.7 & 45.9 \\
& MIKASA~\cite{chang2024mikasa} & 69.7 & 59.4 & 65.4 & 64.0 & 64.4 & 78.6 & 67.3 & \first{70.4} & 75.4 & 75.2 & -- & -- & -- \\
& GPS$^\dagger$~\cite{jia2024sceneverse} & 67.0 & 50.9 & 55.8 & 59.8 & 58.7 & 70.5 & 63.4 & 53.1 & 69.0 & 68.4 & -- & -- & -- \\
& MCLN~\cite{qian2024mcln} & -- & -- & -- & -- & 59.8 & -- & -- & -- & -- & 68.4 & 86.9 & 52.0 & 57.2 \\
& PQ3D$^\dagger$~\cite{zhu2024pq3d} & 73.3 & 56.7 & 60.7 & 67.0 & 64.9 & \second{78.8} & 68.2 & 51.5 & \second{76.7} & 75.6 & 85.2 & 46.8 & 52.8 \\
& LIBA~\cite{wang2025liba} & -- & 57.2 & 60.3 & -- & 64.5 & -- & 70.2 & 61.7 & -- & 75.8 & \second{88.8} & 54.4 & 59.6 \\
& VGMamba~\cite{zhu2025vgmamba} & -- & 61.4 & -- & -- & 68.3 & -- & \first{74.4} & -- & -- & \first{81.3} & \first{91.9} & \second{54.8} & \second{60.0} \\
& ViewSRD~\cite{huang2025viewsrd} & \second{75.3} & \second{64.8} & \second{68.6} & \second{70.6} & \second{69.9} & 78.3 & 70.6 & \second{69.0} & 76.2 & 76.0 & 82.1 & 37.4 & 45.4 \\
& \textbf{Ours} & \first{82.2} & \first{67.2} & \first{70.5} & \first{76.3} & \first{74.6} & \first{81.3} & \second{74.2} & 60.5 & \first{80.0} & \second{79.1} & 84.3 & \first{55.3} & \first{61.0} \\
\midrule
\multirow{7}{*}{\rotatebox{90}{\textbf{Multi Dataset}}}
& 3D-VisTA~\cite{zhu20233d_vista} & 72.1 & 56.7 & 61.5 & 65.1 & 64.2 & 78.8 & 71.3 & 58.9 & 77.3 & 76.4 & 81.6 & 43.7 & 50.6 \\
& GPS~\cite{jia2024sceneverse} & 72.5 & 57.8 & 56.9 & 67.9 & 64.9 & 80.1 & 71.6 & 62.8 & 78.2 & 77.5 & -- & -- & -- \\
& PQ3D~\cite{zhu2024pq3d} & \second{75.0} & \second{58.7} & \second{62.8} & 68.6 & \second{66.7} & \second{82.7} & 72.8 & 62.9 & 80.5 & 79.7 & \second{86.7} & \second{51.5} & 57.0 \\
& Chat-Scene~\cite{huang2024chatscene} & -- & -- & -- & -- & -- & -- & -- & -- & -- & -- & \first{89.6} & 47.8 & 55.5 \\
& LLaVA-3D~\cite{zhu2025llava3d} & -- & -- & -- & -- & -- & -- & -- & -- & -- & -- & -- & -- & 50.1 \\
& UniVLG~\cite{jain2025univlg} & 73.3 & 57.0 & 55.1 & \second{69.9} & 65.2 & \first{84.4} & \first{75.2} & \second{66.2} & \first{82.4} & \first{81.7} & -- & -- & \second{60.7} \\
& \textbf{Ours (\textit{S+R})} & \first{84.1} & \first{68.4} & \first{72.9} & \first{77.5} & \first{76.1} & 82.3 & \second{74.5} & \first{66.6} & \second{80.6} & \second{79.9} & 85.0 & \first{56.4} & \first{62.0} \\
\bottomrule
\end{tabular}
}
\end{table*}

As summarized in \cref{tab:benchmark}, PanoGrounder achieves state-of-the-art or top-2 results on most splits, with clear gains on the human-annotated Nr3D and ScanRefer benchmarks. We observe that mixed training on ScanRefer and ReferIt3D (\emph{S+R}) consistently improves performance, as also seen in 3D-VisTA~\cite{zhu20233d_vista}, GPS~\cite{jia2024sceneverse}, and PQ3D~\cite{zhu2024pq3d} (marked with $^\dagger$ in \cref{tab:benchmark}); we therefore group rows into \emph{Single Dataset} and \emph{Multi Dataset} sections for a fair comparison.

On \textbf{Nr3D}~\cite{achlioptas2020referit3d}, our model achieves the best scores across all splits under single-dataset training, significantly outperforming prior art ViewSRD by +4.7\%. Joint training (S+R) further improves Nr3D to new bests. On \textbf{Sr3D}~\cite{achlioptas2020referit3d}, our method yields Overall 79.1 (single) and 79.9 (S+R), ranking at or near the top across most splits. Notably, our model's stronger gains on Nr3D, which consists of human-generated referring expressions, highlight its robustness in handling diverse, natural language distributions compared to template-based datasets like Sr3D.

On \textbf{ScanRefer}~\cite{chen2020scanrefer}, our model attains the best Overall (61.0) and Multiple (55.3) scores under single-dataset training, outperforming VGMamba (60.0) and LIBA (59.6), with particularly strong gains in multi-object scenes. Joint training yields further improvements to Overall 62.0 and Multiple 56.4, both best. We note that the performance gains on Nr3D are larger than on ScanRefer: ScanRefer's \emph{Unique} subset can often be resolved by simple attribute matching, whereas Nr3D consistently requires fine-grained relational disambiguation, where our stronger language understanding provides the greatest benefit.

\subsubsection{Zero-Shot Comparison with Larger VLMs.}

Recent zero-shot 3DVG methods rely on proprietary or very large VLMs such as GPT-4V and Qwen2-VL-72B. For a fair comparison, we evaluate PanoGrounder without any fine-tuning. As shown in \cref{tab:zeroshot}, our zero-shot pipeline with CogVLM-17B outperforms all baselines despite using a significantly smaller backbone, achieving 53.2 Overall on Nr3D compared to VLM-Grounder's 48.0 with GPT-4V. This suggests that our panoramic multi-modal representation and structured pipeline are effective, enabling strong zero-shot performance even with a smaller backbone.

\begin{table}[t]
  \centering
  \caption{\textbf{Zero-shot comparison on Nr3D.} Without any fine-tuning, PanoGrounder outperforms methods that use significantly larger backbones.}
  \label{tab:zeroshot}
  \begin{tabular}{l l c c c c c}
  \toprule
  \textbf{Method} & \textbf{Backbone} & \textbf{Easy} & \textbf{Hard} & \textbf{VD} & \textbf{VID} & \textbf{Overall} \\
  \midrule
  ZSVG3D~\cite{yuan2024zsvg3d}       & GPT-3.5-turbo  & 46.5 & 31.7 & 36.8 & 40.0 & 39.0 \\
  SeeGround~\cite{li2025seeground}    & Qwen2-VL-72b   & 54.5 & 38.3 & 42.3 & 48.2 & 46.1 \\
  VLM-Grounder~\cite{xu2024vlmgrounder} & GPT-4V         & 55.2 & 39.5 & 45.8 & 49.4 & 48.0 \\
  \midrule
  \textbf{Ours}  & CogVLM-17b     & \first{64.2} & \first{42.7} & \first{49.8} & \first{54.7} & \first{53.2} \\
  \bottomrule
  \end{tabular}
  \end{table}

\begin{figure*}[!t]
    \centering
    \includegraphics[width=\textwidth]{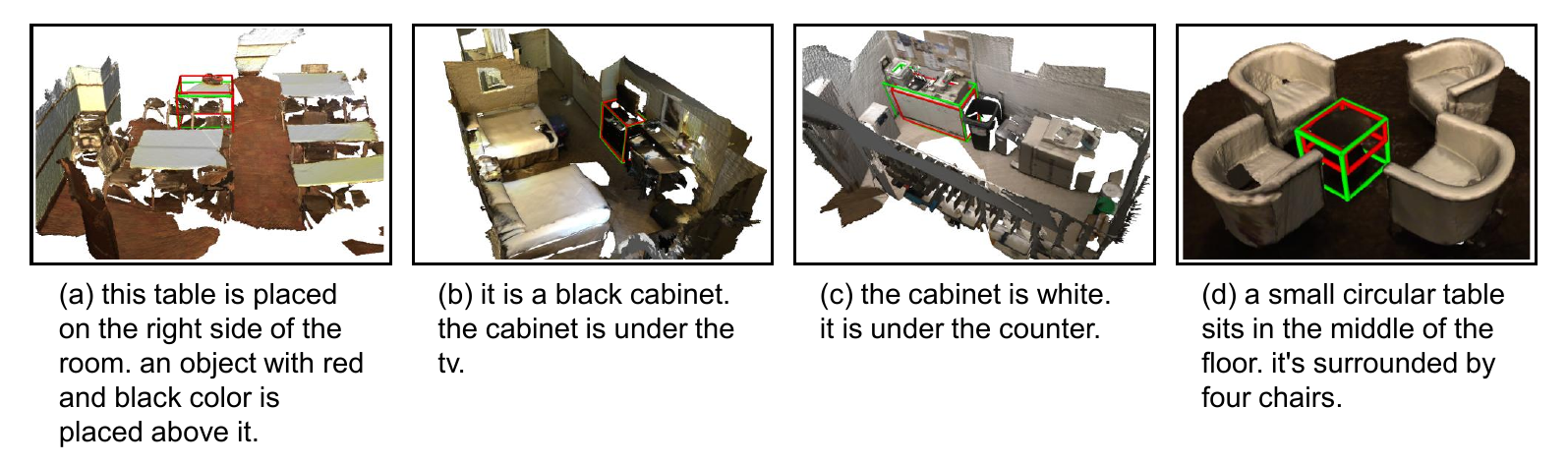}
    \caption{\textbf{Qualitative results on ScanRefer~\cite{chen2020scanrefer}.} 
    Green boxes denote ground-truth objects, and red boxes denote predictions from PanoGrounder. 
    Across diverse scenes and query types, PanoGrounder produces accurate and spatially consistent grounding results.}
    \label{fig:qualitative}
\end{figure*}

\subsubsection{Qualitative Results}

\cref{fig:qualitative} shows qualitative grounding results on ScanRefer~\cite{chen2020scanrefer}. PanoGrounder accurately localizes the referred objects and produces spatially consistent 3D boxes across diverse scenes and query types. In \cref{fig:qualitative}(a), PanoGrounder correctly distinguishes the target table from nearby distractors using relational cues. Similarly, in \cref{fig:qualitative}(b--d), it resolves challenging references in cluttered scenes by combining attribute cues with spatial and relational reasoning. We also observe three common failure modes---ambiguous references satisfied by multiple objects, small or heavily occluded targets, and partial localization of multi-part objects---and provide representative examples and a detailed analysis in the supplementary material.

\begin{table*}[t]
\centering
\caption{\textbf{Cross-dataset evaluation on ScanNetV2~\cite{dai2017scannet}, ARKitScenes~\cite{dehghan2021arkitscenes}, and 3RScan~\cite{wald2019rio}.} All models are trained on ScanRefer~\cite{chen2020scanrefer} and directly evaluated on unseen 3D scenes from ARKitScenes and 3RScan with corresponding text annotations. To ensure fair comparison focused on 3D scene generalization, we use GT object segmentation for all methods; * additionally leverages ground-truth semantic labels. All baseline results are re-run by us.}
\label{tab:unseen_dataset}
\resizebox{\textwidth}{!}{%
\small
\begin{tabular}{lccc ccc ccc}
\toprule
\multirow{2.5}{*}{\textbf{Method}} & \multicolumn{3}{c}{\textbf{ScanRefer}} & \multicolumn{3}{c}{\textbf{ARKitScenes}} & \multicolumn{3}{c}{\textbf{3RScan}} \\
\cmidrule(lr){2-4} \cmidrule(lr){5-7} \cmidrule(lr){8-10}
 & unique & multiple & overall & unique & multiple & overall & unique & multiple & overall \\
\midrule
ViL3DRel~\cite{chen2022vil3drel}    & 92.0 & 51.8 & 59.6 & 57.2 & 21.1 & 28.3 & 71.8 & 31.3 & 36.8 \\
3D-VisTA~\cite{zhu20233d_vista}    & 89.5 & 49.9 & 57.2 & 59.7 & 26.6 & 32.9 & \second{74.1} & \second{32.0} & \second{37.7} \\
BUTD-DETR*~\cite{jain2022butddetr}  & \second{92.5} & 52.6 & 58.5 & \second{66.3} & \second{30.6} & \second{36.1} & -- & -- & -- \\
MCLN*~\cite{qian2024mcln}       & \first{93.4} & \second{54.9} & \second{60.6} & 61.2 & 30.0 & 35.3 & -- & -- & -- \\
\midrule
\textbf{Ours}     & 91.7 & \first{58.5} & \first{64.9} & \first{74.2} & \first{48.0} & \first{53.5} & \first{80.4} & \first{37.7} & \first{43.8} \\
\bottomrule
\end{tabular}%
}
\end{table*}

\subsection{Generalization Analysis}
\label{sec:generalization}

To fairly evaluate scene and language generalization independently of proposal quality, we fix proposals to ground-truth object segmentations and re-run all baselines under the same setting in the following experiments.

\subsubsection{Scene Generalization.}

We evaluate scene-level generalization by testing on two unseen 3D datasets: ARKitScenes~\cite{dehghan2021arkitscenes} paired with human-written expressions from SceneVerse~\cite{jia2024sceneverse}, and 3RScan~\cite{wald2019rio} paired with RIORefer~\cite{miyanishi2024cross3dvg}. As shown in \cref{tab:unseen_dataset}, PanoGrounder shows notably smaller performance degradation than existing baselines when transferred to unseen scenes. This indicates that our approach generalizes effectively to novel environments and diverse 3D layouts beyond the training distribution. BUTD-DETR~\cite{jain2022butddetr} and MCLN~\cite{qian2024mcln} could not be evaluated on 3RScan+RIORefer because their public implementations are restricted to the ScanNetV2 object set.

\subsubsection{Text Generalization.}
\label{sec:textgeneralization}

We assess the linguistic robustness of 3D visual grounding models on ScanRefer~\cite{chen2020scanrefer} by generating four variants of each query with LLaMA~3.3~\cite{dubey2024llama}: \textbf{Para.} (paraphrased), \textbf{+Aff.} (affordance/functional description added), \textbf{+Aff.$-$N} (\textbf{+Aff.} with target noun removed), and \textbf{Mask} (target noun replaced with ``object'').

As shown in \cref{tab:text_aug}, PanoGrounder performs on par with PQ3D on \textbf{Org.} and \textbf{Para.}, and surpasses it on \textbf{+Aff.} (+2.7). The gains become particularly pronounced once explicit class cues are removed, with improvements of +15.9 on \textbf{+Aff.$-$N} and +8.5 on \textbf{Mask}. In these two variants, where the target object name is absent, the model must reason over spatial and contextual cues rather than simple lexical matching. PanoGrounder remains accurate under such implicit or underspecified language, demonstrating strong reasoning and robust performance across diverse textual perturbations.

\subsection{Ablation Study}

\cref{tab:ablation} analyzes how input modalities and auxiliary objectives contribute to performance. The EMD loss consistently improves accuracy regardless of input configuration: (A)$\rightarrow$(B) (57.5$\rightarrow$58.4) and (D)$\rightarrow$(E) (59.3$\rightarrow$60.2), confirming its value as a distance-aware regularization signal. Incorporating semantic features in (C) further boosts accuracy to 60.4, indicating that multi-view fused semantic context is highly beneficial even without geometric input. Row (F) shows that geometric QA without geometric input (60.7) also falls short of the full model. Our final configuration (\textbf{Ours}), which combines RGB, semantic, and geometric features with both EMD and geometric QA, achieves the best accuracy (61.0). Notably, masking the geometric features of \textbf{Ours} at inference drops its accuracy to 60.5, confirming their utility. Together, these results suggest that geometric input and geometric QA act synergistically rather than as interchangeable components.

\begin{table*}[!t]
\begin{minipage}[t]{0.51\linewidth}
    \centering
\centering
\caption{\textbf{Robustness to ScanRefer~\cite{chen2020scanrefer} text variants.} 
Top-1 accuracy on original queries and four text modifications (see text for details). 
All models use GT object segmentation; * additionally uses GT semantic labels. All baseline results are re-run by us.}
\renewcommand{\arraystretch}{1.2}
\resizebox{\linewidth}{!}{
\begin{tabular}{lccccc}
\toprule
\textbf{Method} & \textbf{Org.} & \textbf{Para.} & \textbf{+Aff.} & \textbf{+Aff.$-$N} & \textbf{Mask} \\
\midrule
ViL3DRel~\cite{chen2022vil3drel} & 59.6 & 41.5 & 9.7 & 6.5 & 22.9 \\
3D-VisTA~\cite{zhu20233d_vista} & 57.2 & 52.9 & 50.4 & 28.3 & 28.5 \\
BUTD-DETR*~\cite{jain2022butddetr} & 58.5 & 53.4 & 55.1 & 35.5 & 27.3 \\
MCLN*~\cite{qian2024mcln} & 60.6 & 54.6 & 55.7 & 35.7 & 31.5 \\
PQ3D~\cite{zhu2024pq3d} & \first{66.0} & \first{63.3} & \second{58.6} & \second{35.8} & \second{33.3} \\
\midrule
\textbf{Ours} & \second{64.9} & \second{61.5} & \first{61.3} & \first{51.7} & \first{41.8} \\
\textbf{$\Delta$ (Ours $-$ PQ3D)}
& \negd{-1.1}
& \negd{-1.8}
& \posd{+2.7}
& \posd{+15.9}
& \posd{+8.5} \\
\bottomrule
\end{tabular}
}
\label{tab:text_aug}
\end{minipage}
\hfill
\begin{minipage}[t]{0.45\linewidth}
    \centering
\centering
\caption{\textbf{Ablation study.} We evaluate the contribution of each input modality (RGB, semantic, geometric) and auxiliary training signals (EMD loss and geometric QA).}
\label{tab:ablation}
\resizebox{\linewidth}{!}{
\begin{tabular}{c ccccc c}
\toprule
& \multicolumn{3}{c}{\textbf{Input}} & 
\multicolumn{2}{c}{\textbf{Train}} & 
\multirow{2.5}{*}{\textbf{Acc@0.25}} \\ 
\cmidrule(lr){2-4} \cmidrule(lr){5-6}
& RGB & Sem. & Geo. & 
EMD & Geo QA & \\ 
\midrule
(A) & \cmark & \xmark & \xmark & \xmark & \xmark  & 57.5 \\
(B) & \cmark & \xmark & \xmark & \cmark & \xmark  & 58.4 \\
(C) & \cmark & \cmark & \xmark & \cmark & \xmark  & 60.4 \\
(D) & \cmark & \cmark & \cmark & \xmark & \xmark  & 59.3 \\
(E) & \cmark & \cmark & \cmark & \cmark & \xmark  & 60.2 \\
(F) & \cmark & \cmark & \xmark & \cmark & \cmark  & 60.7 \\
\midrule
\textbf{Ours} & \cmark & \cmark & \cmark & \cmark & \cmark & \first{61.0} \\
\bottomrule
\end{tabular}
}

\end{minipage}
\end{table*}

\begin{table*}[t]
\centering
\caption{\textbf{Comparison of panoramic and pinhole view settings.} 
We report accuracy by view configuration (rows) and by the number of objects referenced in the text (columns). Our panoramic representation substantially outperforms all pinhole baselines without oracle knowledge, and the performance gap widens as the query mentions more surrounding objects. The oracle GT target setting provides an upper bound for pinhole-based approaches.}
\label{tab:panorama}
\resizebox{\linewidth}{!}{%
\small
\begin{tabular}{lcccccc}
\toprule
\multirow{2.5}{*}{\textbf{View setting}} & \multirow{2.5}{*}{\textbf{Overall}} &
\multicolumn{5}{c}{\textbf{\# of objects mentioned in the text}} \\
\cmidrule(lr){3-7}
 & & 0 (0.6\%) & 1 (23.7\%) & 2 (46.6\%) & 3 (21.9\%) & 4+ (7.2\%) \\
\midrule
pinhole @ 4 views                      & 41.3 & 41.1 & 38.7 & 42.1 & 42.0 & 42.2 \\
pinhole @ 16 views                     & 51.0 & 50.0 & 47.3 & 52.4 & 51.0 & 54.5 \\
pinhole @ Semantic~\cite{li2025seeground} & 43.6 & 37.5 & 42.3 & 44.3 & 43.6 & 44.0 \\
\textbf{Ours}                          & \first{61.0} & \first{49.9} & \first{54.7} & \first{61.3} & \first{64.3} & \first{71.0} \\
\cmidrule(lr){1-7}
pinhole @ GT target & 67.5 & 58.9 & 65.5 & 69.3 & 66.5 & 66.0 \\
\bottomrule
\end{tabular}%
}
\end{table*}

\subsubsection{Panorama vs. Pinhole Cameras.}

Panoramic views offer two primary advantages over pinhole cameras: (i) they eliminate the need for viewing direction prediction, thereby simplifying the pipeline; and (ii) they capture the maximum number of text-referenced objects within a single view, strengthening context and relation modeling.

We compare our panoramic representation against several pinhole-based view selection strategies, all using a $120^\circ$ horizontal FoV and a fixed camera center as in \cref{sec:scp}.
(1) \textbf{Fixed rotation (4 / 16 views):} cameras are rotated at fixed horizontal angular intervals to uniformly cover $360^\circ$.
(2) \textbf{Semantic-based view selection (4 views):} following SeeGround~\cite{li2025seeground}, instance masks and predicted labels of Mask3D~\cite{Schult23mask3d} are used to find text-mentioned objects, up to four of which are chosen by projected size on the panorama, and the camera is oriented toward each; if fewer than four objects are mentioned, the remaining directions are sampled randomly.
(3) \textbf{GT target (1 view, oracle):} a pinhole camera that always points directly at the ground-truth target.

As shown in \cref{tab:panorama}, panoramic views substantially outperform all non-oracle pinhole baselines (41.3--51.0 vs.\ 61.0 overall), indicating that our panoramic representation captures scene context more effectively. The gap further widens as the number of referenced objects increases: in the ``4+ objects'' category, our approach (71.0) even surpasses the pinhole oracle (66.0). This suggests that pinhole views are fundamentally limited by their narrow field of view, which observes only a subset of objects at once, whereas panoramic views preserve the global spatial relations of the scene.
\section{Conclusion}

The proposed 3DVG method, PanoGrounder, has demonstrated that panoramic renderings are an effective intermediate representation for 3D visual grounding, preserving long-range spatial relations within a single view while enabling direct use of powerful 2D VLMs. Extensive experiments also have shown state-of-the-art results on ScanRefer and Nr3D, and strong robustness to unseen scenes and text rephrasings, indicating that panorama-driven VLM grounding is a practical path toward generalizable 3D visual grounding.

\subsubsection{Limitations and Future Work.}
PanoGrounder requires an off-the-shelf 3D reconstruction as a preprocessing step, and severe reconstruction noise and artifacts can degrade 2D inference. It is also slower than fully feed-forward 3D methods due to VLM processing, though this can be mitigated by using smaller VLM backbones. Looking forward, three promising directions are: (i) handling queries with \emph{no target object} or \emph{multiple targets}; (ii) generalizing PanoGrounder to \emph{multi-task} settings such as 3D captioning and 3D VQA; and (iii) extending it to building-level or outdoor settings via more flexible camera placement beyond the floor-based grid.

\section*{Acknowledgments}
We thank Chunghyun Park for his helpful comments and technical advice during the development of the model. This work was partly supported by Institute of Information \& communications Technology Planning \& Evaluation (IITP) grant funded by the Korea government (MSIT) [NO.RS-2021-II211343, Artificial Intelligence Graduate School Program (Seoul National University)] and National Research Foundation of Korea (NRF) grant funded by the Korea government (MSIT) [No. RS-2024-00359718].

%
%
\bibliographystyle{splncs04}
\bibliography{main}

\clearpage
\def\thefootnote{\arabic{footnote}}
\title{PanoGrounder: Bridging 2D and 3D with Panoramic Scene Representations \\ for VLM-based 3D Visual Grounding \\ (Supplementary Material)}
\titlerunning{PanoGrounder}
\author{Seongmin Jung\inst{1}$^{\star}$\orcidlink{0009-0002-5281-4375} \and
Seongho Choi\inst{1,2}$^{\star}$\orcidlink{0009-0007-4458-6823} \and
Gunwoo Jeon\inst{1}\orcidlink{0009-0008-1410-9556} \and
Minsu Cho\inst{3}\orcidlink{0000-0001-7030-1958} \and \\
Jongwoo Lim\inst{1}\orcidlink{0000-0002-2814-4765}}
\authorrunning{S.~Jung et al.}
\institute{Seoul National University, South Korea\\
\email{\{sm18570, sh.choi, gunw0, jongwoo.lim\}@snu.ac.kr}\and
Robotics Lab, Hyundai Motor Company, South Korea\and
Pohang University of Science and Technology (POSTECH), South Korea\\
\email{mscho@postech.ac.kr}\\
\href{https://choiseongho-h.github.io/PanoGrounder}{https://choiseongho-h.github.io/PanoGrounder}}
\maketitle
\let\thefootnote\relax\footnotetext{$^{\star}$ Equal contribution.}

\appendix

\section{Implementation Details}

\subsection{Camera Placement and Rendering Stage}
\label{sec:supp_impl_rendering}

In this section, we further elaborate on how we select panoramic viewpoints across scenes and render color, range, and feature maps from each viewpoint.

\subsubsection{Floor estimation and candidate grid.}
We estimate the floor plane using RANSAC~\cite{fischler1981ransac}. Specifically, we downsample the point cloud (mesh vertices or 3DGS centers) to a $5\,\mathrm{cm}$ resolution, retain the bottom $50\%$ of points based on height as floor candidates, and run $10{,}000$ RANSAC iterations. At each iteration, we sample three points to fit a plane, enforce a nearly vertical normal ($|n_z| > 0.9$), and count inliers within a threshold of $10^{-3}\,\mathrm{m}$, ultimately selecting the plane with the highest inlier count.

Subsequently, we compute a 2D convex hull over all scene points to obtain a tight floor footprint and create a 2D grid over this hull with a spacing of $h{=}0.1\,\mathrm{m}$. The $z$-coordinates of the grid centers are set to the average height of the raw camera poses.

\subsection{Multi-Modal Panorama}
\label{sec:supp_mmpr}

We provide specific implementation details for constructing the multi-modal panoramic representations described in Sec. 3.2.

\paragraph{Rendering Configuration and Dimensions.}
All panoramic inputs are generated using Equirectangular Projection (ERP) with a $360^\circ$ horizontal and $180^\circ$ vertical field of view.
We differentiate the rendering resolution based on the modality:
\begin{itemize}
    \item \textbf{RGB and Depth:} Rendered at $(H, W) = (490, 490)$. These are subsequently processed with a patch size of $14 \times 14$, resulting in a flattened token sequence of length $N_f = 1225$ (corresponding to a $35 \times 35$ grid).
    \item \textbf{Semantic Features:} Rendered directly at the target resolution of $35 \times 35$ to align with the token grid without further patching.
\end{itemize}
Regarding the underlying 3D representations, we utilize PyTorch3D for mesh-based scenes and ODGS~\cite{lee2024odgs} for 3D Gaussian Splatting (3DGS) scenes.

\paragraph{Feature Extraction Details.}
The geometric and semantic feature maps are constructed as follows:
\begin{itemize}
    \item \textbf{Geometric Features ($\mathbf{f}_{\mathrm{geo}}$):} The rendered depth map $D_k \in \mathbb{R}^{490 \times 490}$ is first clipped and min-max normalized to $[0, 1]$. This normalized map is fed into a lightweight ViT encoder (trainable DINOv2~\cite{oquab2023dinov2}) with a patch size of 14, outputting $\mathbf{f}_{\mathrm{geo}} \in \mathbb{R}^{N_f \times d_{\mathrm{geo}}}$ ($d_{\mathrm{geo}} = 384$).
    
    \item \textbf{Semantic Features ($\mathbf{f}_{\mathrm{sem}}$):} We first extract raw feature maps from a frozen ViT encoder applied to the original RGB views. These raw features are bilinearly interpolated to the resolution of the original RGB images. The upsampled features are then lifted to 3D: for mesh scenes, we project per-pixel features onto vertices using camera parameters; for 3DGS scenes, we adopt~\cite{yue2024fit3d} to lift features onto Gaussians. Finally, the aggregated 3D semantic field is rendered directly at the $35 \times 35$ resolution, yielding $\mathbf{f}_{\mathrm{sem}} \in \mathbb{R}^{N_f \times d_{\mathrm{sem}}}$ ($d_{\mathrm{sem}} = 384$).
\end{itemize}

\subsection{Prompt Design}

\begin{figure*}[p]
    \centering

    \begin{tcolorbox}[
        width=\textwidth,
        colback=gray!5,
        colframe=gray!40,
        coltitle=black,
        title=(a) Prompt used for querying PanoGrounder
    ]
    \scriptsize\ttfamily
    Please locate the object described as: <description>.\\
    If there are multiple, pick the most prominent.\\
    Provide the bounding box coordinates as [x1,y1,x2,y2]. Output only the coordinates.
    \end{tcolorbox}


    \begin{tcolorbox}[
        width=\textwidth,
        colback=gray!5,
        colframe=gray!40,
        coltitle=black,
        title=(b) Prompt for generating paraphrased queries (Para.)
    ]
    \scriptsize\ttfamily
    You are a precise paraphraser.\\
    Rewrite the given sentence so that the meaning is exactly the same, but the wording and phrasing are different.\\
    Do NOT add or remove any information. Keep entities, counts, colors, and spatial relations unchanged.\\
    <description>.
    \end{tcolorbox}


    \begin{tcolorbox}[
        width=\textwidth,
        colback=gray!5,
        colframe=gray!40,
        coltitle=black,
        title=(c) Prompt for generating affordance-based queries (+Aff.)
    ]
    \scriptsize\ttfamily
    You are an affordance-based rewriter.\\
    Goal: Describe what the speaker wants to DO with the implied target object.\\
    Use first-person language ("I want to …", "I need to …", "I'd like to …").\\
    Keep ALL constraints (attributes like color/size, counts, spatial relations such as left/right/behind/between/next to/in front of/inside, and references to nearby items) EXACTLY the same; you may reorder or use synonyms. Do NOT add or remove information.\\
    IMPORTANT: Pick an action that is TYPICAL for the correct answer word (object affordance). AVOID generic verbs unless they are the prototypical affordance for that object type.\\
    <description>.
    \end{tcolorbox}


    \begin{tcolorbox}[
        width=\textwidth,
        colback=gray!5,
        colframe=gray!40,
        coltitle=black,
        title=(d) Prompt for generating affordance-based queries without target noun (+Aff.$-$N)
    ]
    \scriptsize\ttfamily
    You are an affordance-based rewriter.\\
    Goal: Describe what the speaker wants to DO with the implied target object, WITHOUT ever naming that object or any of its synonyms.\\
    Use first-person language ("I want to …", "I need to …", "I'd like to …").\\
    Keep ALL constraints (attributes like color/size, counts, spatial relations such as left/right/behind/between/next to/in front of/inside, and references to nearby items) EXACTLY the same; you may reorder or use synonyms. Do NOT add or remove information.\\
    IMPORTANT: Pick an action that is TYPICAL for the correct answer word (object affordance). AVOID generic verbs unless they are the prototypical affordance for that object type. NEVER name the target object or any near-synonyms.\\
    <description>.
    \end{tcolorbox}

    \caption{\textbf{Prompt templates for VLM inference and text augmentation.}
    (a) Instruction-style prompt for querying PanoGrounder with a panoramic image and a referring expression \texttt{<description>}.
    (b) LLaMA~3.3 prompt for paraphrasing queries (Para.\@ in Tab. 4).
    (c) LLaMA~3.3 prompt for generating affordance-focused descriptions that retain the object class name (+Aff.\@ in Tab. 4).
    (d) LLaMA~3.3 prompt for generating affordance-focused descriptions that exclude the target noun (+Aff.$-$N in Tab. 4).}
    \label{fig:prompt}
\end{figure*}

To clearly define the role of the VLM and enforce a consistent output format, we encapsulate the original referring expressions from the dataset within a concise instruction-style prompt. \cref{fig:prompt}(a) illustrates the specific prompt template employed in the ScanRefer experiments. Furthermore, to evaluate text generalization (Sec. 4.3), we leverage LLaMA~3.3~\cite{dubey2024llama} to generate linguistic variants of the original queries, specifically focusing on rephrasing and affordance-based descriptions. We designed three distinct instruction prompts for these text modifications, as depicted in \cref{fig:prompt}(b)--(d).

\begin{itemize}
    \item \textbf{Masking (Mask).}
    We substitute instances of the ground-truth class label in the text with a generic placeholder ``\texttt{object}'', while preserving the remainder of the sentence structure. This is a deterministic string replacement that does not involve an LLM.

    \item \textbf{Paraphrasing (Para.).}
    Utilizing the instruction prompt in \cref{fig:prompt}(b), LLaMA~3.3 paraphrases the sentence to alter its syntax and vocabulary, ensuring that the semantic context and constraints associated with the target object remain intact.

    \item \textbf{Affordance (+Aff.).}
    Guided by the prompt in \cref{fig:prompt}(c), LLaMA~3.3 generates a first-person description emphasizing the intended interaction with the target object (i.e., affordance), while maintaining all original constraints. The generated query retains the target object's class name.

    \item \textbf{Affordance without target noun (+Aff.$-$N).}
    Using the prompt in \cref{fig:prompt}(d), the model follows a similar affordance-focused instruction but is explicitly restricted from naming the target object's class label or its near-synonyms. Consequently, the query relies exclusively on functional descriptions and contextual cues.
\end{itemize}

\subsection{VLM Hyperparameters}

In this section, we detail the hyperparameters used for training and inference of the VLM.
We fine-tune the backbone using Low-Rank Adaptation (LoRA) with a rank of $r=64$ and $\alpha=64$.
The model is trained for 5 epochs on the ScanRefer dataset with a batch size of 64, utilizing the Adam optimizer with a learning rate of $1 \times 10^{-4}$.
Because multiple cameras are deployed per scene, we count these epochs scene-centrically: all panoramic views of a scene constitute a single pass. Since each referring expression is observed from approximately 2.4 viewpoints on average, one scene-centric epoch entails about $2.4\times$ more iterations than a standard text-centric epoch on ScanRefer or ReferIt3D, so our 5 scene-centric epochs correspond to roughly 12 text-centric epochs.
During training, if the ground-truth object is not visible in a specific panoramic image, that sample is excluded from the batch.
For the supervision signal, the Earth Mover's Distance (EMD) loss is calculated based on coordinates normalized by the maximum output image coordinate (999), and its weight is set to $\lambda = 10.0$.

Furthermore, we incorporate auxiliary Geometric QA samples into the training pipeline, which constitute $1/3$ of the total data.
For each sample, we randomly select two pixels on a panorama, provide their coordinates in $[x_1,y_1,x_2,y_2]$ format, and ask the model which point has a larger coordinate value along a given axis ($x$, $y$, or $z$). The model answers with the coordinates of the selected point.
These auxiliary samples require no human annotation and are generated on-the-fly during data loading, encouraging the model to ground the injected geometric features into spatial reasoning through language.

\section{Two-Stage Variant}

Two-stage methods follow a proposal-and-selection strategy. In the ReferIt3D~\cite{achlioptas2020referit3d} benchmark, the task inherently provides candidate ground-truth (GT) point cloud segments. Similarly, for the generalization experiments (Sec. 4.3), to ensure a fair comparison against existing baselines~\cite{zhu20233d_vista, chang2024mikasa, zhu2024pq3d} that rely on varying segmentors, we utilize GT masks for all methods.

The inference pipeline operates as follows: The 2D VLM first processes the equirectangular panorama to predict a 2D bounding box. We then match this prediction to a specific 3D instance by computing the IoU with the projected GT instance masks, assigning the prediction to the instance with the highest IoU.

Finally, we aggregate these frame-level predictions into a single decision. For ScanRefer, we utilize standard majority voting. For ReferIt3D, leveraging the known candidate geometries, we employ a weighted voting scheme based on the visible proportion of each candidate in the frame, thereby favoring viewpoints where the target is more clearly observable.

\section{Additional Analysis and Ablations}

In this section, we provide additional ablation studies to further evaluate the effectiveness of various components of our method. Unless otherwise noted, all ablation studies are conducted on ScanRefer~\cite{chen2020scanrefer} and report Acc@0.25.

\subsection{Scene Representation}

We evaluate two scene representations: a triangle mesh and a 3D Gaussian Splatting (3DGS) model~\cite{kerbl20233dgs}, training each variant on panoramas rendered from its respective representation. For Ours (3DGS), we initialize the Gaussians from the ground-truth mesh vertices and train with the ground-truth camera trajectories. As shown in \cref{tab:mesh3dgs}, the mesh-based representation outperforms the 3DGS model across all metrics, as the densification process during 3DGS training can introduce floater artifacts in under-reconstructed regions, degrading panoramic rendering quality. Nevertheless, our 3DGS-based model still surpasses LIFT-GS~\cite{cao2025liftgs}, a recent 3D visual grounding method built upon 3DGS.

We additionally evaluate a fully automated setting (3DGS* in \cref{tab:mesh3dgs}), where the 3D scene is reconstructed entirely from raw RGB images via COLMAP~\cite{schoenberger2016sfm, schoenberger2016mvs} and 3DGS, without any manual mesh cleaning or ground-truth camera trajectories. Unlike the mesh representation, which benefits from human cleaning that removes under-reconstructed regions and artifacts, this variant relies on a fully automated pipeline and is therefore more susceptible to reconstruction noise. This variant uses the same model checkpoint as Ours (3DGS)---only the panoramas used at evaluation time are rendered from the automatically reconstructed scene. Since COLMAP fails to register sufficient images on some ScanNet~\cite{dai2017scannet} scenes, we exclude scenes where fewer than 300 images are registered, leaving 92 out of 141 validation scenes. For a fair comparison, all three of our variants are evaluated on this same subset, while the LIFT-GS result is cited from the original paper. This demonstrates practical viability when only raw RGB images are available.

\begin{figure*}[p]
    \centering

    \begin{minipage}[c]{0.04\textwidth}
        \centering
        \rotatebox[origin=c]{90}{\texttt{scene0000\_00}}
    \end{minipage}%
    \begin{minipage}[c]{0.48\textwidth}
        \centering
        \textbf{Mesh renderings}\\[1mm]  
        \includegraphics[width=0.98\linewidth]{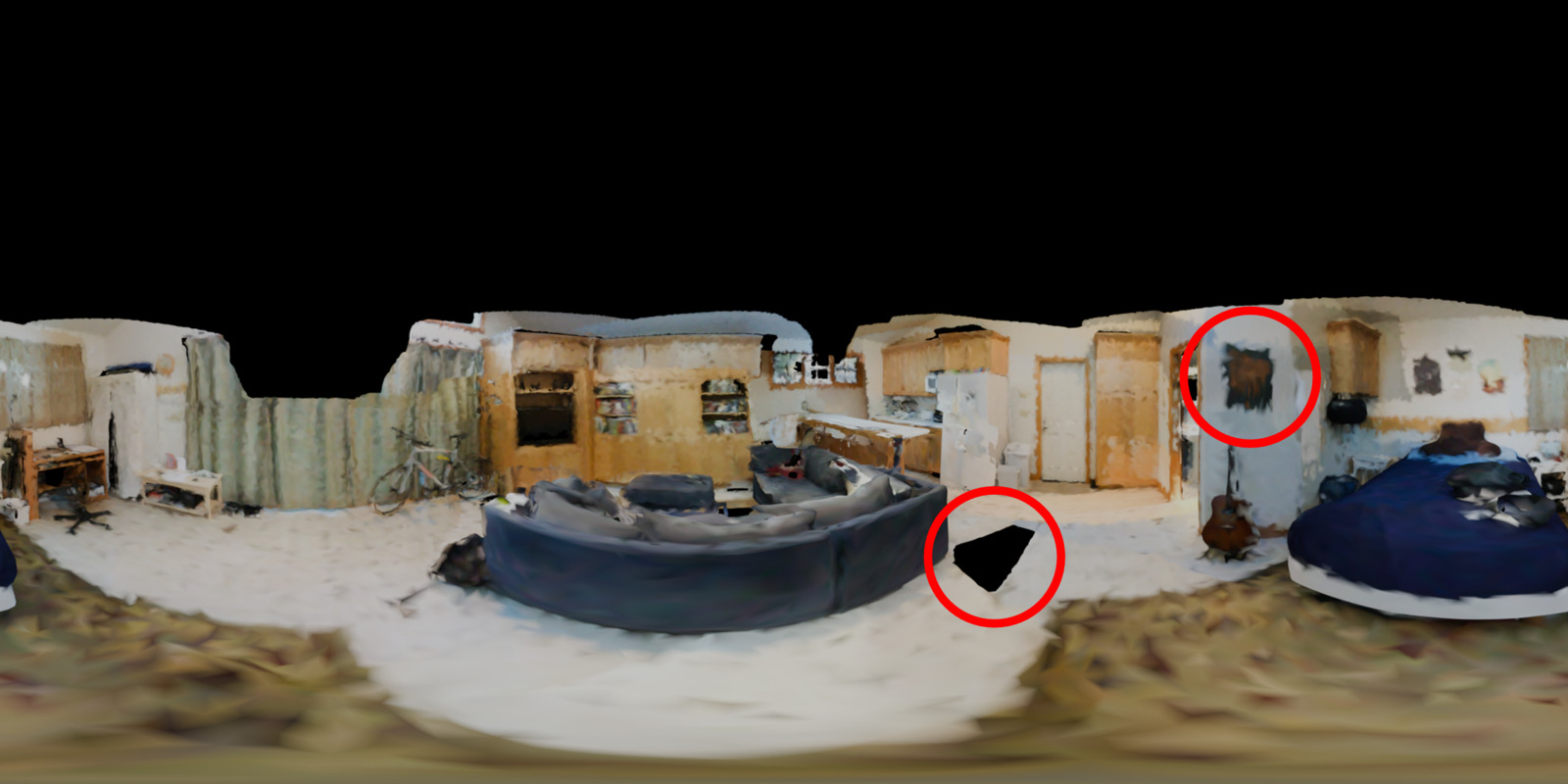}
    \end{minipage}%
    \begin{minipage}[c]{0.48\textwidth}
        \centering
        \textbf{3DGS renderings}\\[1mm]  
        \includegraphics[width=0.98\linewidth]{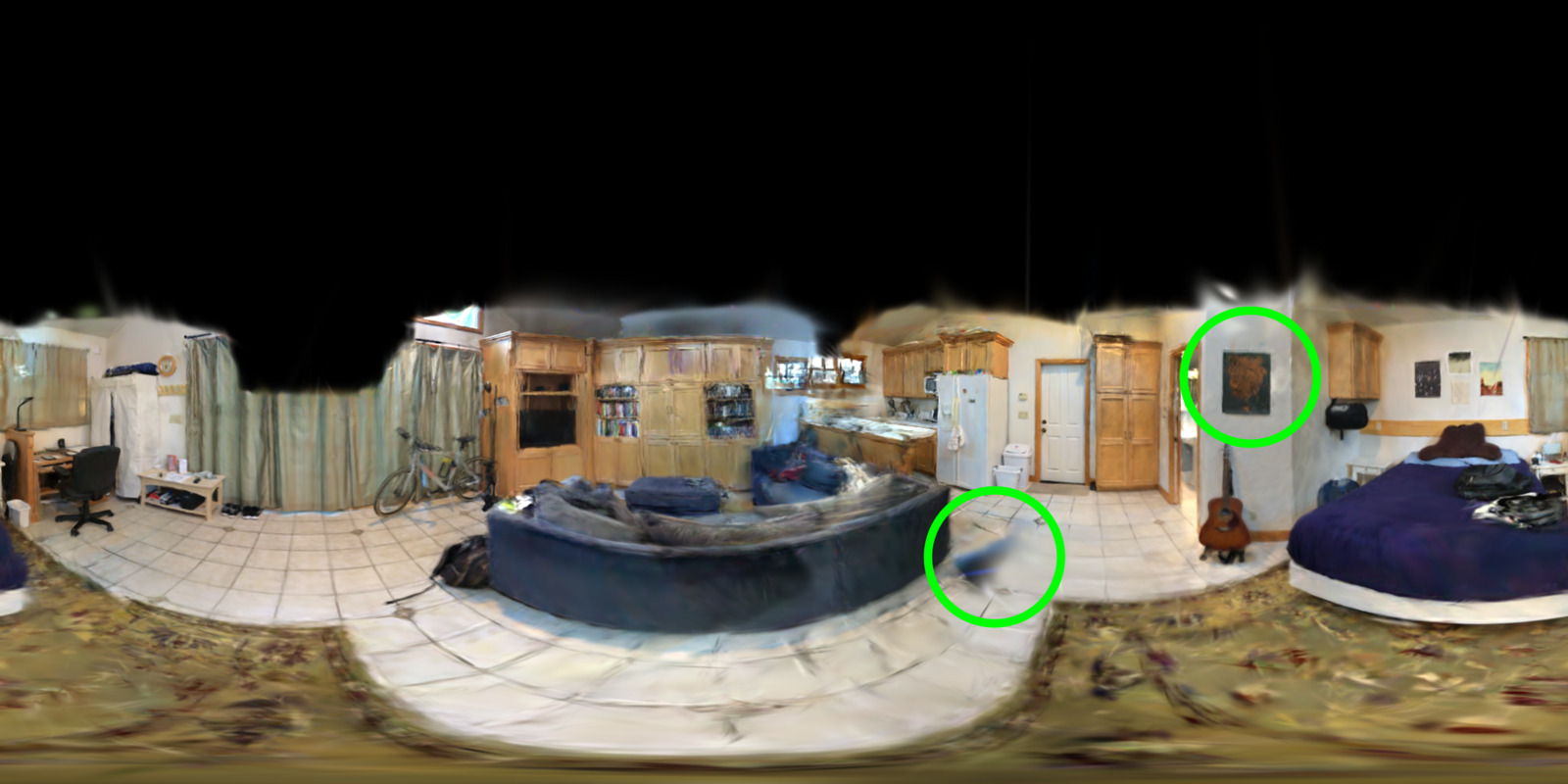}
    \end{minipage}

    \vspace{1mm}

    \begin{minipage}[c]{0.04\textwidth}
        \centering
        \rotatebox[origin=c]{90}{\texttt{scene0300\_00}}
    \end{minipage}%
    \begin{minipage}[c]{0.48\textwidth}
        \centering
        \includegraphics[width=0.98\linewidth]{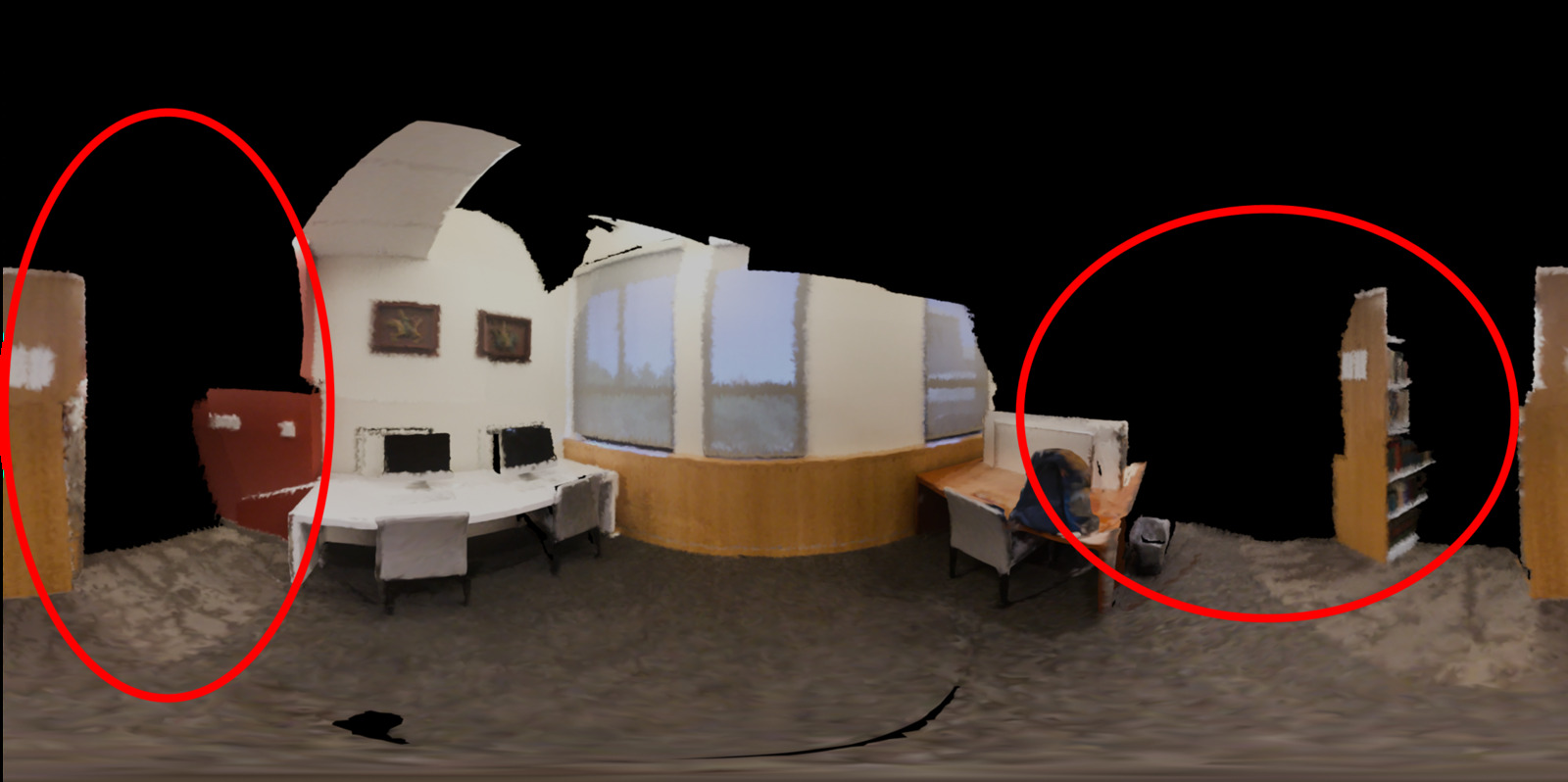}
    \end{minipage}%
    \begin{minipage}[c]{0.48\textwidth}
        \centering
        \includegraphics[width=0.98\linewidth]{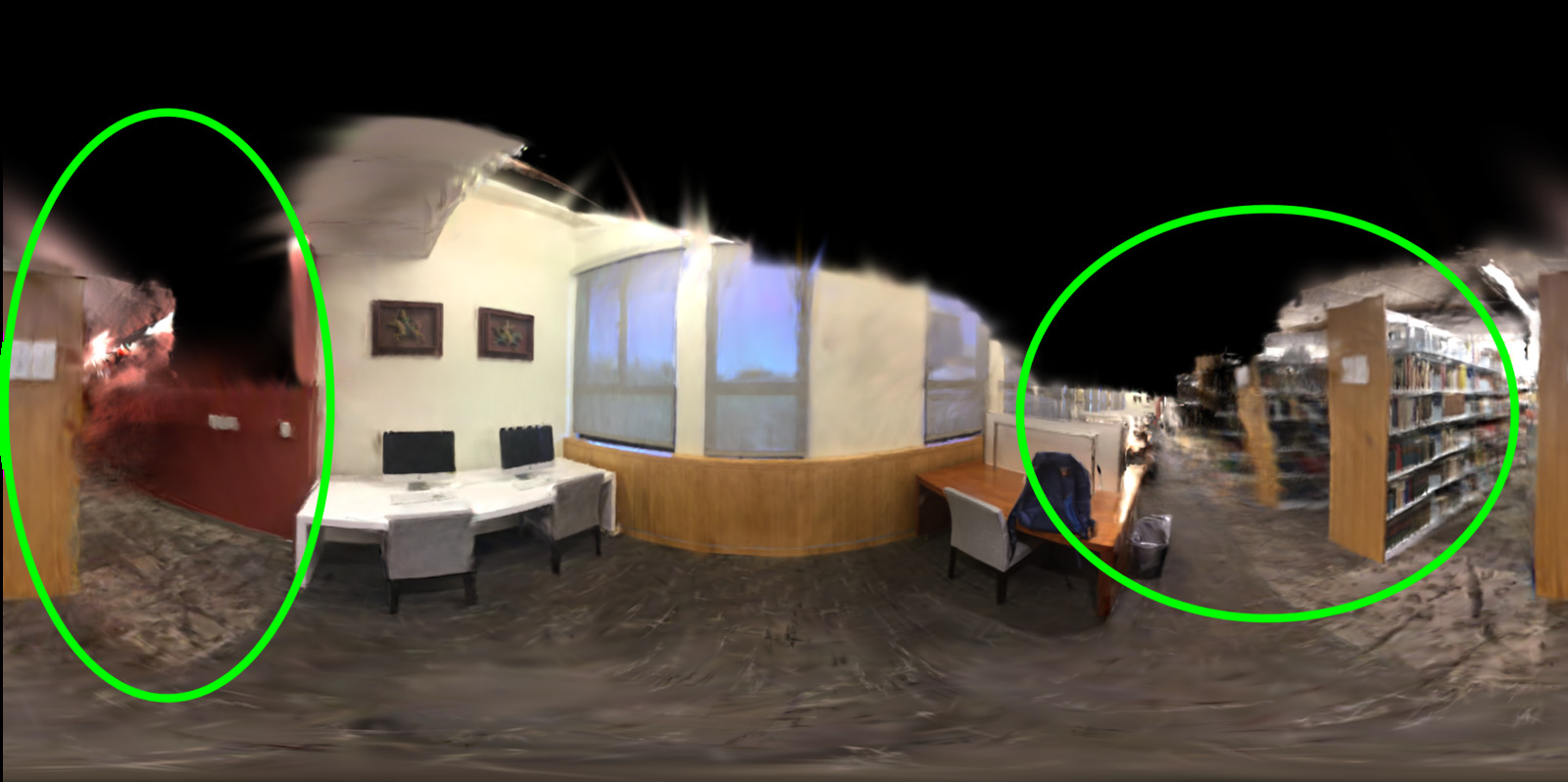}
    \end{minipage}

    \vspace{1mm}

    \begin{minipage}[c]{0.04\textwidth}
        \centering
        \rotatebox[origin=c]{90}{\texttt{scene0008\_00}}
    \end{minipage}%
    \begin{minipage}[c]{0.48\textwidth}
        \centering
        \includegraphics[width=0.98\linewidth]{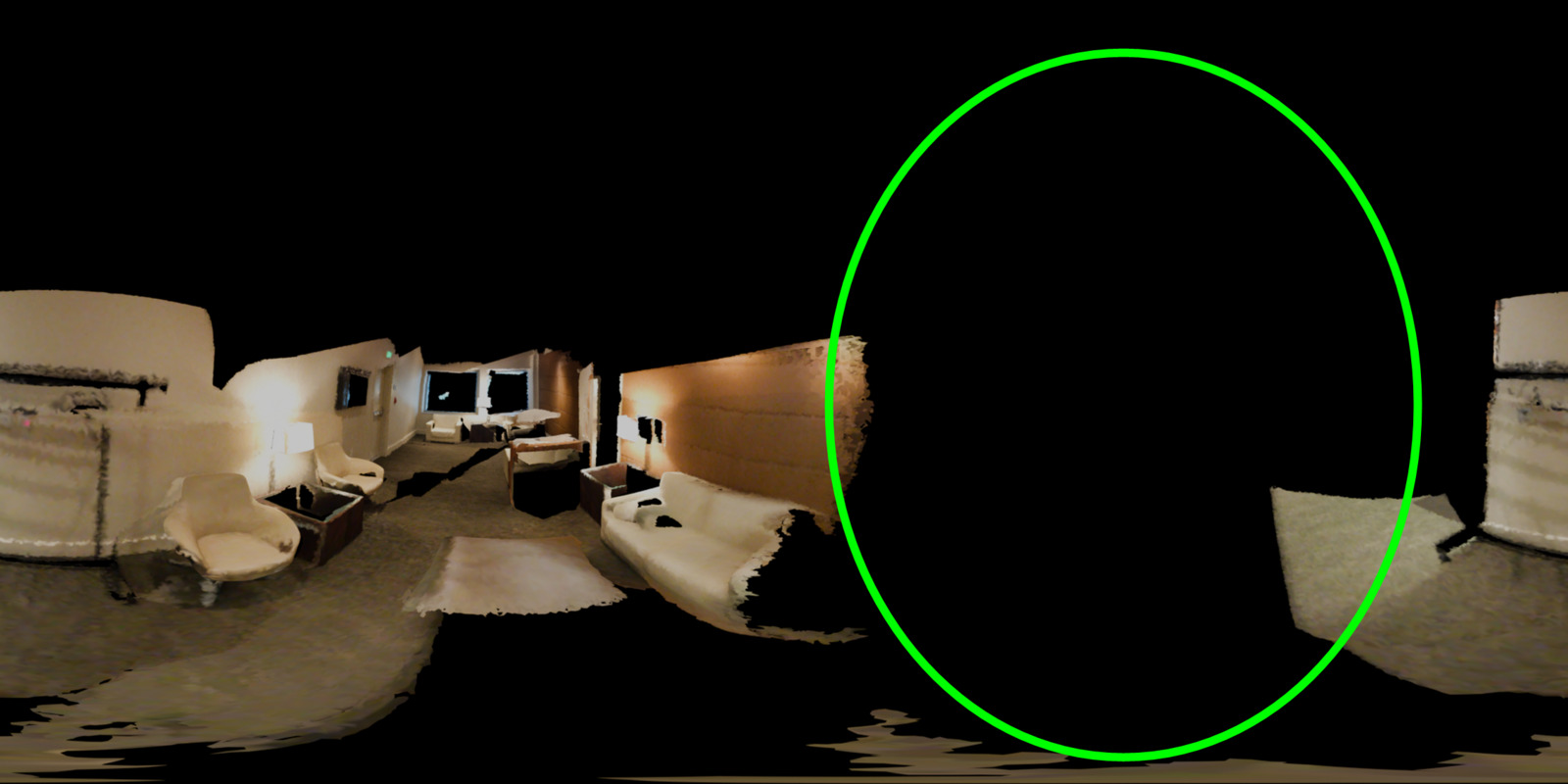}
    \end{minipage}%
    \begin{minipage}[c]{0.48\textwidth}
        \centering
        \includegraphics[width=0.98\linewidth]{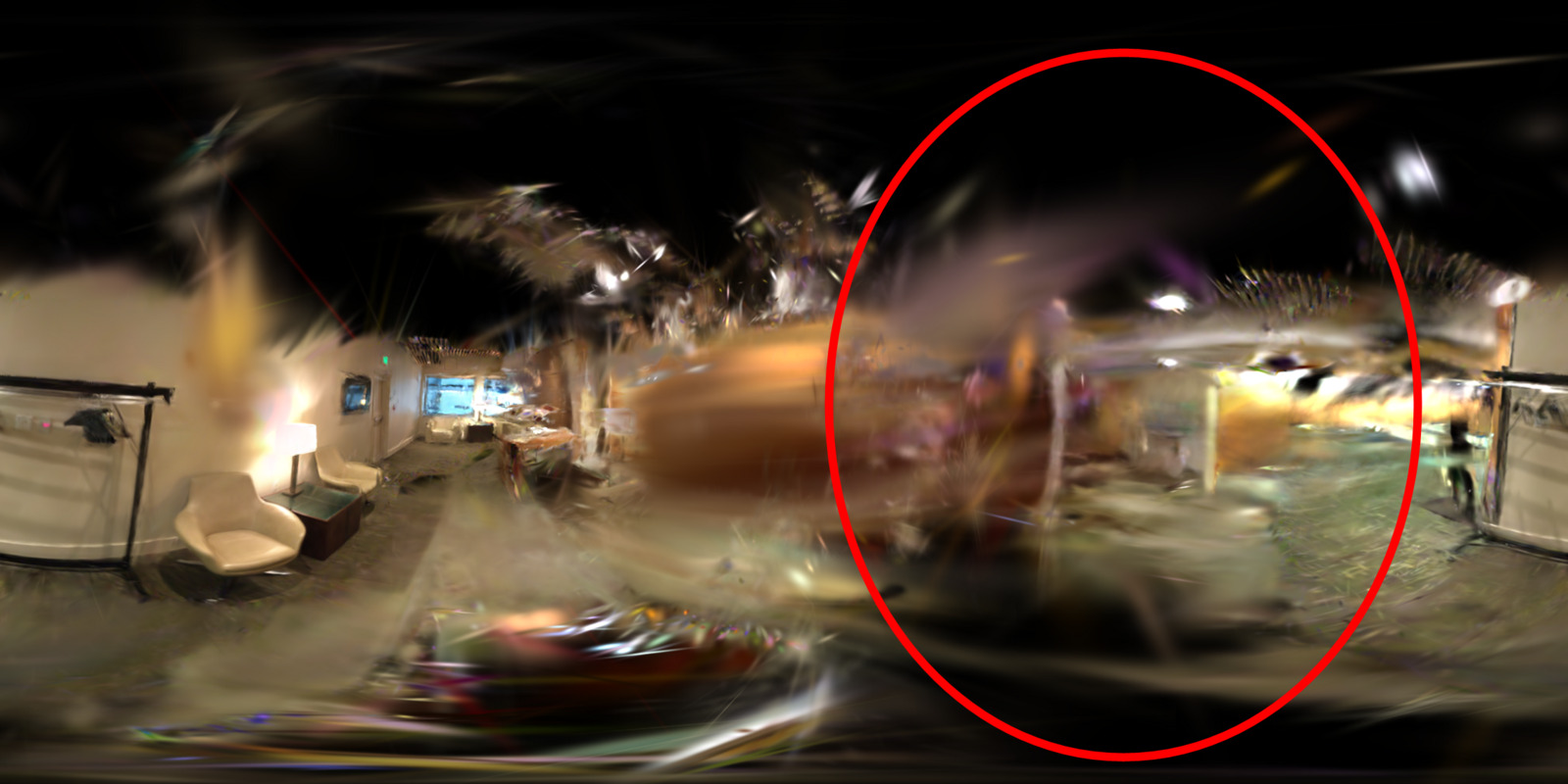}
    \end{minipage}

    \vspace{1mm}

    \begin{minipage}[c]{0.04\textwidth}
        \centering
        \rotatebox[origin=c]{90}{\texttt{scene0623\_00}}
    \end{minipage}%
    \begin{minipage}[c]{0.48\textwidth}
        \centering
        \includegraphics[width=0.98\linewidth]{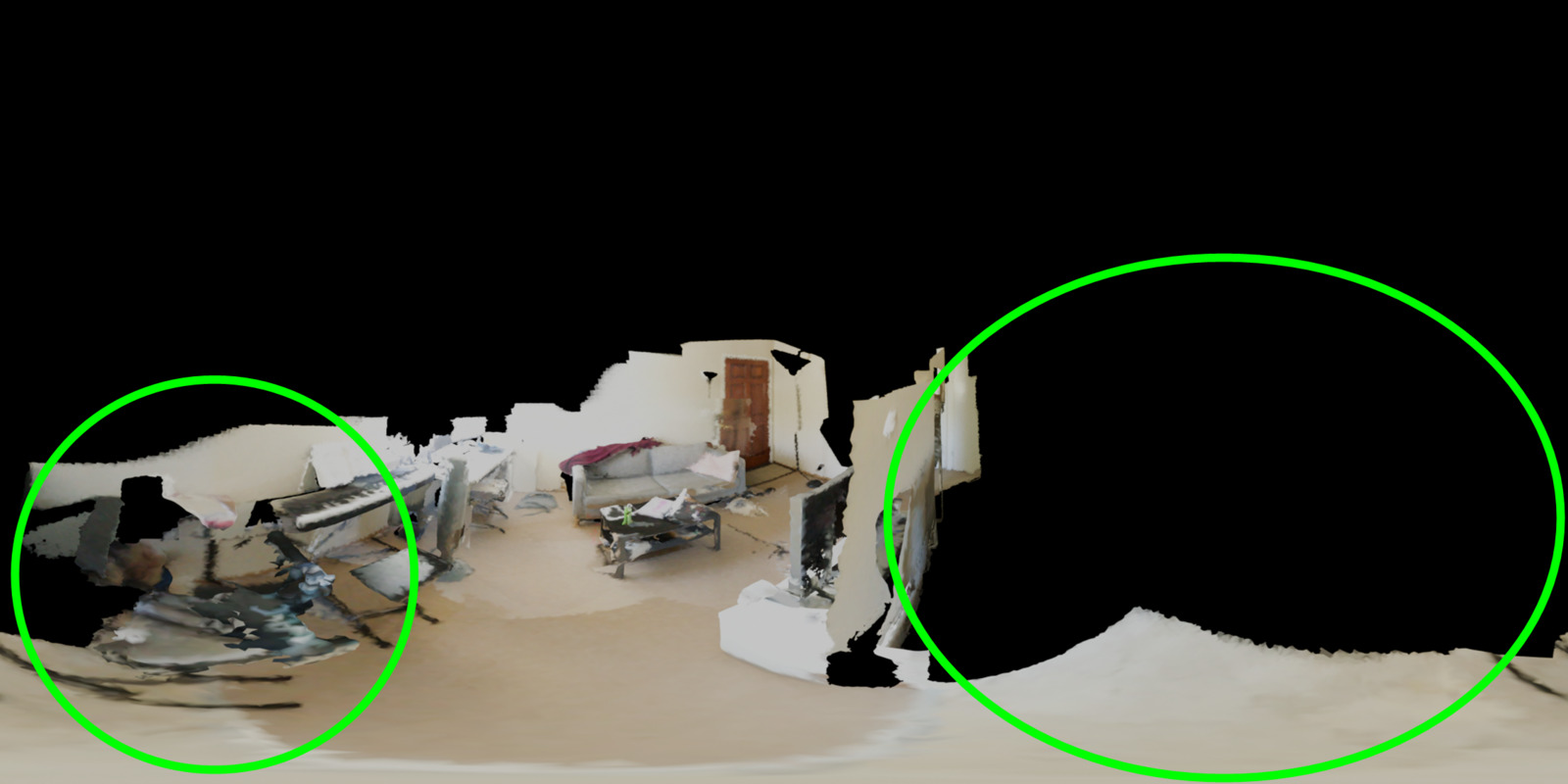}
    \end{minipage}%
    \begin{minipage}[c]{0.48\textwidth}
        \centering
        \includegraphics[width=0.98\linewidth]{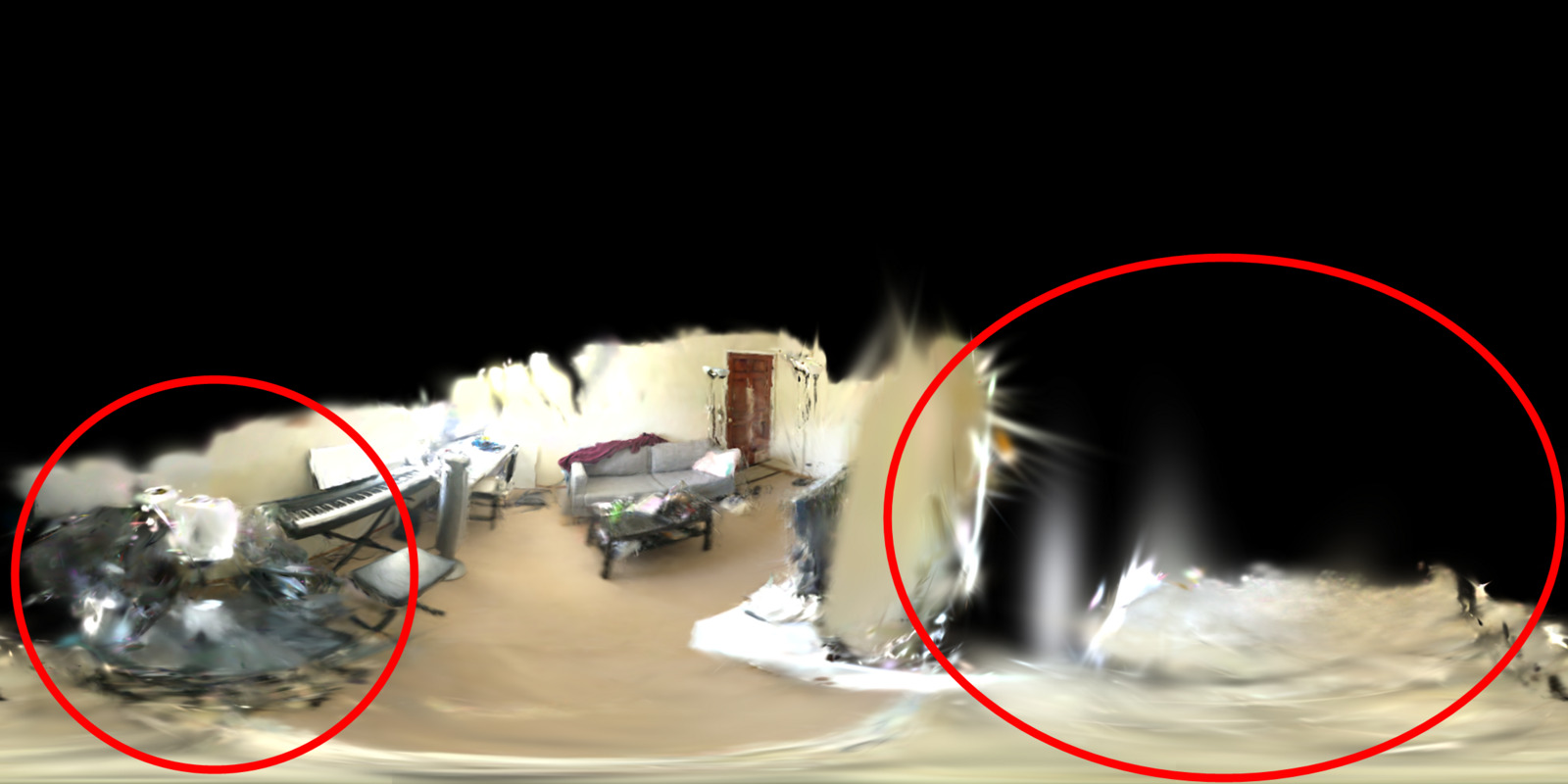}
    \end{minipage}

    \caption{\textbf{Mesh vs 3DGS panoramic renderings across different scenes.} Green circles mark well-reconstructed regions and red circles mark degraded ones (holes or floater artifacts). 3DGS fills holes left by the mesh and sharpens boundaries in the top two scenes, but introduces floater artifacts in under-reconstructed regions of the bottom two.}
    \label{fig:mesh3dgs_scenes}
\end{figure*}

\cref{fig:mesh3dgs_scenes} illustrates qualitative differences between the two representations. In scenes such as \texttt{scene0000\_00}, 3DGS demonstrates superior surface continuity (e.g., on the floor) and sharper object boundaries (e.g., pictures on the wall). Similarly, in \texttt{scene0300\_00}, 3DGS successfully reconstructs distant areas that the mesh fails to capture. Conversely, in scenes with incomplete reconstructions (e.g., \texttt{scene0008\_00} and \texttt{scene0623\_00}), 3DGS is prone to artifacts, which degrades rendering quality.

\begin{table}[p]
\centering
\caption{\textbf{Impact of scene representation.} Acc@0.25 on ScanRefer~\cite{chen2020scanrefer} with different scene representations. * denotes a fully automated reconstruction without ground-truth meshes or camera trajectories.}
\label{tab:mesh3dgs}
\begin{tabular}{lccc}
\toprule
\textbf{Method} & \textbf{Unique} & \textbf{Multiple} & \textbf{Overall} \\
\midrule
LIFT-GS~\cite{cao2025liftgs} & -- & -- & 49.7 \\
Ours (3DGS*) & 73.1 & 50.6 & 55.1 \\
Ours (3DGS) & 80.7 & 55.8 & 60.8 \\
Ours (Mesh) & \textbf{82.7} & \textbf{56.3} & \textbf{61.6} \\
\bottomrule
\end{tabular}
\end{table}

\subsection{Structure-aware Camera Placement}

\begin{table}[t]
\centering
\caption{\textbf{Ablation on camera placement strategy.} We report the average number of cameras per scene and Acc@0.25 for each scoring factor combination. The \textit{Random} baseline uses the same number of cameras as our method.}
\label{table:camera_placement}
\begin{tabular}{cccccc}
\toprule
& \textbf{RC} & \textbf{DS} & \textbf{DT} & 
\textbf{\#Cam/Scene} &
\textbf{Acc@0.25} \\ 
\midrule
Random     & \xmark & \xmark & \xmark & 2.4 & 51.9 \\
(a)     & \cmark & \xmark & \xmark & 2.5 & 55.7 \\
(b)     & \cmark & \cmark & \xmark & 2.4 & 59.2 \\
\textbf{Ours} & \cmark & \cmark & \cmark & \textbf{2.4} & \textbf{61.0} \\
\bottomrule
\end{tabular}
\end{table}

\begin{figure}[!t]
    \centering
    \includegraphics[width=0.6\columnwidth]{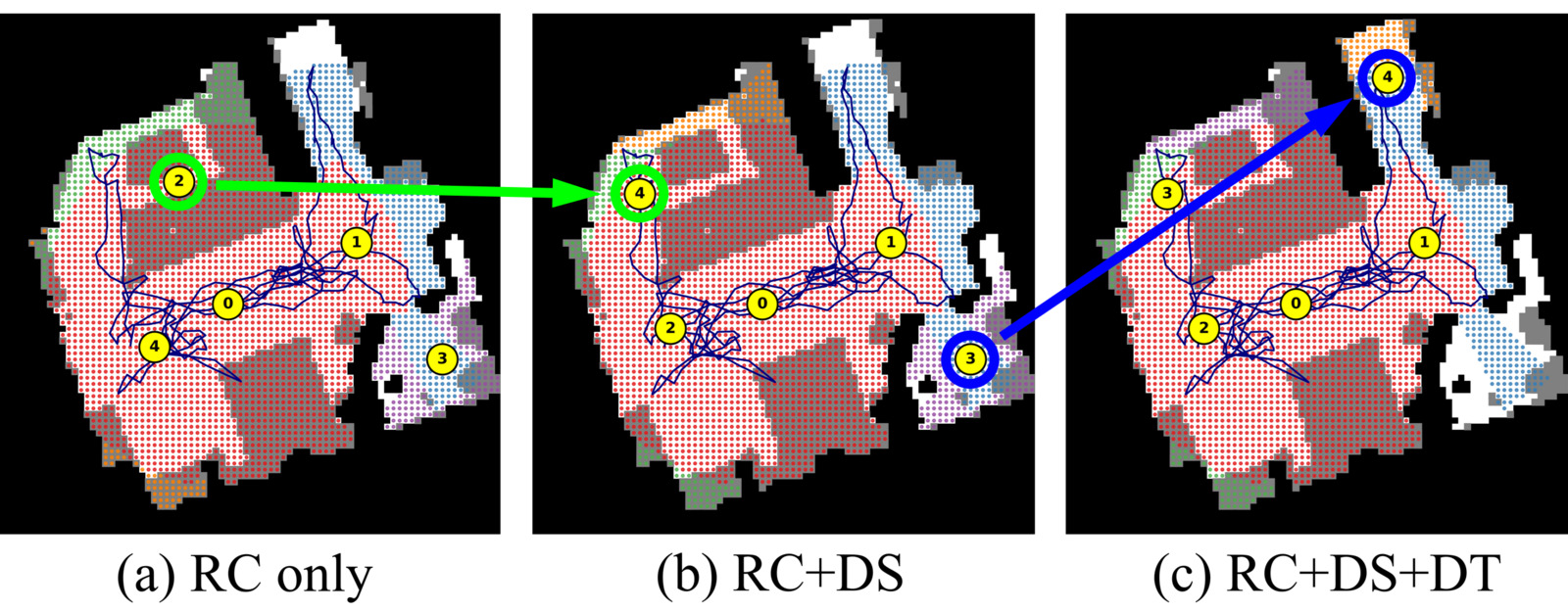}
    \caption{\textbf{Effect of camera placement factors.} Top-down views of a scene where white cells show the floor layout, gray cells denote obstacles, and the blue curve depicts the raw RGB camera trajectory. Yellow numbered markers indicate selected panoramic cameras, and the colored dots around each camera visualize the ray coverage contributed by that camera. Comparing the green circles in (a) and (b), adding distance-to-surface (RC+DS) moves cameras away from obstacles into more open regions. Comparing the blue circles in (b) and (c), further adding distance-to-trajectory (RC+DS+DT) keeps cameras near the original capture path while still achieving good coverage.}
    \label{fig:camera_placement}
\end{figure}

\cref{table:camera_placement} and \cref{fig:camera_placement} present an ablation study of the scoring factors employed in our \textit{structure-aware camera placement} module. While using only ray coverage (RC) improves upon random selection, cameras may still be positioned too close to obstacles. In \cref{fig:camera_placement}(a), the green circle marks a viewpoint located directly on top of the sofa in the top-left region. By incorporating the distance-to-surface factor (RC+DS), as shown in (b), the corresponding viewpoint (green circle) is shifted away from the sofa and slightly outward. This adjustment prevents distortion near the bottom of the panorama and yields a significant accuracy gain (55.7 $\rightarrow$ 59.2). However, RC+DS may still place cameras in geometrically incomplete areas; the blue circle in (b) highlights a viewpoint inside the bathroom---a region the ground-truth trajectory never enters---resulting in incomplete reconstruction. Our full strategy (RC+DS+DT) additionally incorporates the distance-to-trajectory factor to discourage placement in such out-of-trajectory regions. Consequently, the blue circle in (c) shows this viewpoint shifted back toward the main path near the kitchen, effectively filling the remaining uncovered space. This configuration achieves optimal scene coverage and rendering quality, reaching the highest grounding accuracy with a comparable number of cameras per scene.

\begin{figure}[!t]
    \centering
    \includegraphics[width=0.5\columnwidth]{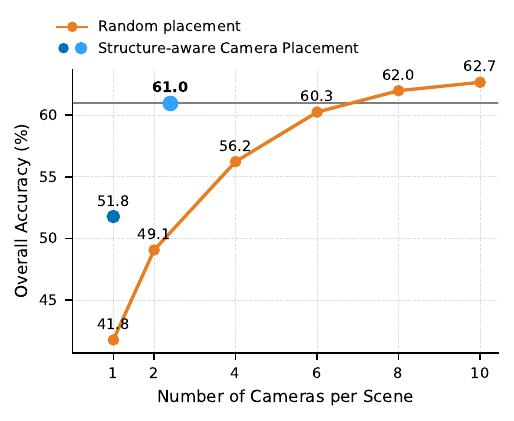}
    \caption{\textbf{Effect of the number of cameras per scene on overall accuracy.} The orange curve shows random placement, whose performance monotonically increases as more cameras are placed in the scene. Blue markers denote our Structure-aware Camera Placement: the left marker corresponds to the best single-camera configuration, and the right marker corresponds to our full model with an average of only 2.4 cameras per scene, which achieves overall accuracy roughly equivalent to using about 7 randomly placed cameras.}
    \label{fig:cam_place_abl}
\end{figure}

\cref{fig:cam_place_abl} illustrates the effect of the number of cameras per scene. With random placement, overall accuracy increases monotonically as more cameras are added, though the incremental gain diminishes as the scene becomes saturated with views. Our Structure-aware Camera Placement provides two key reference points (blue markers): (i) the best single-view, which selects the single camera with the highest score and significantly outperforms the random single-view baseline; and (ii) the full-selection, averaging 2.4 cameras per scene, which achieves an overall accuracy comparable to that of approximately seven randomly placed cameras. This demonstrates that our strategy substantially improves both overall accuracy and computational efficiency.

\begin{figure}[!t]
    \centering
    \includegraphics[width=\columnwidth]{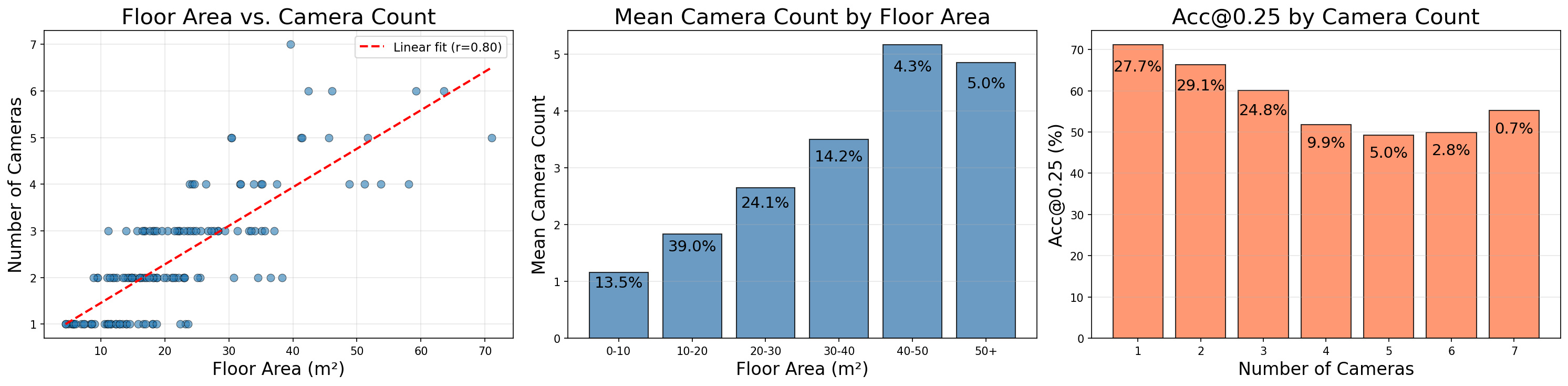}
    \caption{\textbf{Relationship between floor area and camera count.} (Left) Scatter plot of floor area vs.\ the number of cameras selected by our structure-aware placement, with a linear fit ($r{=}0.80$). (Center) Mean camera count per floor area bin; percentages indicate the proportion of scenes in each bin. (Right) Acc@0.25 as a function of the number of cameras; percentages indicate the proportion of scenes with that camera count. Our placement strategy adaptively assigns more cameras to larger scenes, maintaining consistent coverage across varying room sizes.}
    \label{fig:floor_area_vs_cams}
\end{figure}

\cref{fig:floor_area_vs_cams} further analyzes the relationship between scene size and the number of cameras selected by our placement strategy. The floor area (computed from the convex hull of the scene point cloud) exhibits a strong positive correlation ($r{=}0.80$) with the number of selected cameras, confirming that our strategy adaptively allocates more viewpoints to larger scenes. The right panel shows that accuracy remains relatively stable across different camera counts, suggesting that the placement strategy effectively maintains sufficient coverage regardless of scene scale.

\begin{figure}[t]
    \centering
    \begin{minipage}[c]{0.46\linewidth}
        \centering
        \includegraphics[width=\linewidth]{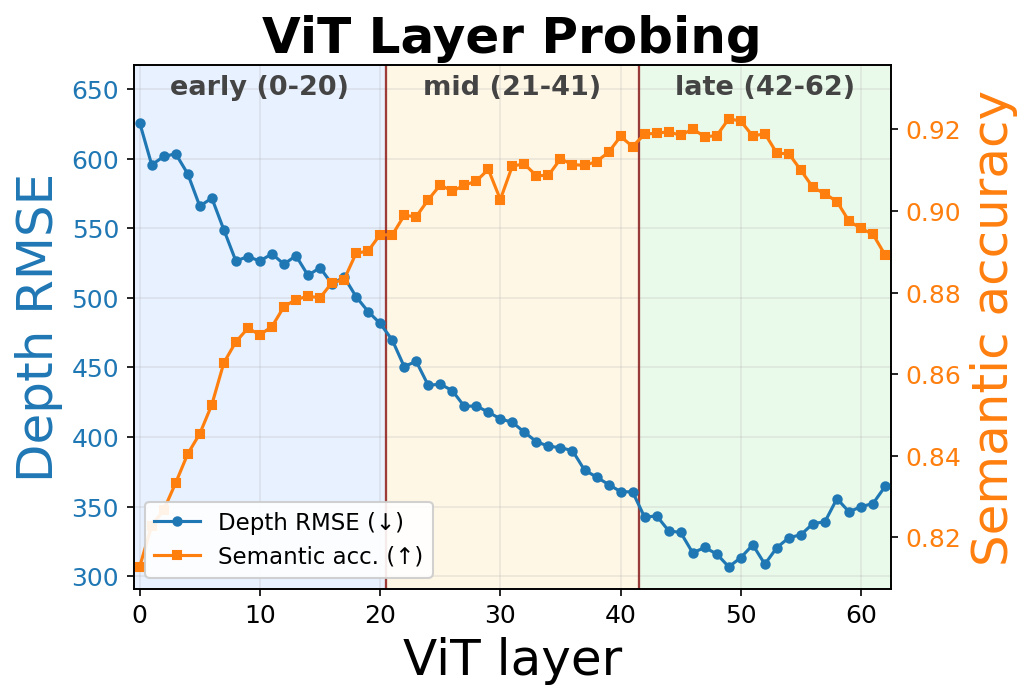}
        \captionof{figure}{\textbf{Probing CogVLM's frozen ViT.} Per-layer performance of the linear probes for depth and object label.}
        \label{fig:probe}
    \end{minipage}\hfill
    \begin{minipage}[c]{0.48\linewidth}
        \centering
        \captionof{table}{\textbf{Adapter layer placement on ScanRefer.} Acc@0.25 for alternative geometric/semantic injection bands.}
        \label{tab:layer_placement}
        \medskip\smallskip
        \renewcommand{\arraystretch}{1.2}
        \begin{tabular*}{\linewidth}{@{\extracolsep{\fill}}lccc@{}}
        \toprule
        \textbf{Config} & \textbf{Geo} & \textbf{Sem} & \textbf{Acc@0.25} \\
        \midrule
        Both-mid      & mid           & mid           & 61.0 \\
        Both-late     & late          & late          & 60.9 \\
        Swap          & late          & mid           & 60.9 \\
        \textbf{Ours} & \textbf{mid}  & \textbf{late} & \first{61.0} \\
        \bottomrule
        \end{tabular*}
    \end{minipage}
\end{figure}

\subsection{Adapter Layer Placement}

Our design injects geometric features into mid-level layers and semantic features into later layers (Sec.~3.2). We hypothesize that injecting each feature where the backbone already represents the corresponding information aids its integration. To identify these layers, we probe what CogVLM's frozen ViT already encodes (\cref{fig:probe}). At each layer, we fit two linear probes mapping each patch token to (i)~its per-patch depth (ridge regression) and (ii)~its dominant object label (logistic regression). Both probes perform poorly in the first few layers and improve sharply thereafter, showing that geometric and semantic information is captured progressively and concentrated in the mid-to-late layers. We therefore place geometric cues at mid layers and semantic cues at late layers, following the well-documented tendency of deep vision encoders to capture low-level geometry before high-level semantics~\cite{zeiler2014visualizing, raghu2021vitseelikecnn}.

To verify this choice, we move each adapter between the mid and late layers while holding the training setup fixed. As \cref{tab:layer_placement} shows, the resulting configurations span only 60.9--61.0\% Acc@0.25, indicating that the exact placement is not critical; we adopt geometric-mid and semantic-late as one effective choice.

\subsection{Adapter Initialization}

We investigate the impact of using a zero-initialized convolution layer within the feature adapter. This ablation study was conducted using only the multi-view semantic feature encoder, excluding the geometry encoder and Geometric QA. We observe that replacing the zero-initialization of the $1 \times 1$ convolution layer with Gaussian initialization degrades performance to 59.5\%, falling short of the baseline adapter configuration which achieves 60.4\%. The baseline configuration, combining a 2-layer MLP with a zero-initialized $1 \times 1$ convolution, highlights the importance of this initialization strategy. Zero initialization helps preserve the pretrained VLM behavior while enabling the adapter to effectively learn task-specific context.

\subsection{Training Convergence}

\begin{figure*}[!t]
    \centering
    \begin{subfigure}[t]{0.48\textwidth}
        \centering
        \includegraphics[width=\textwidth]{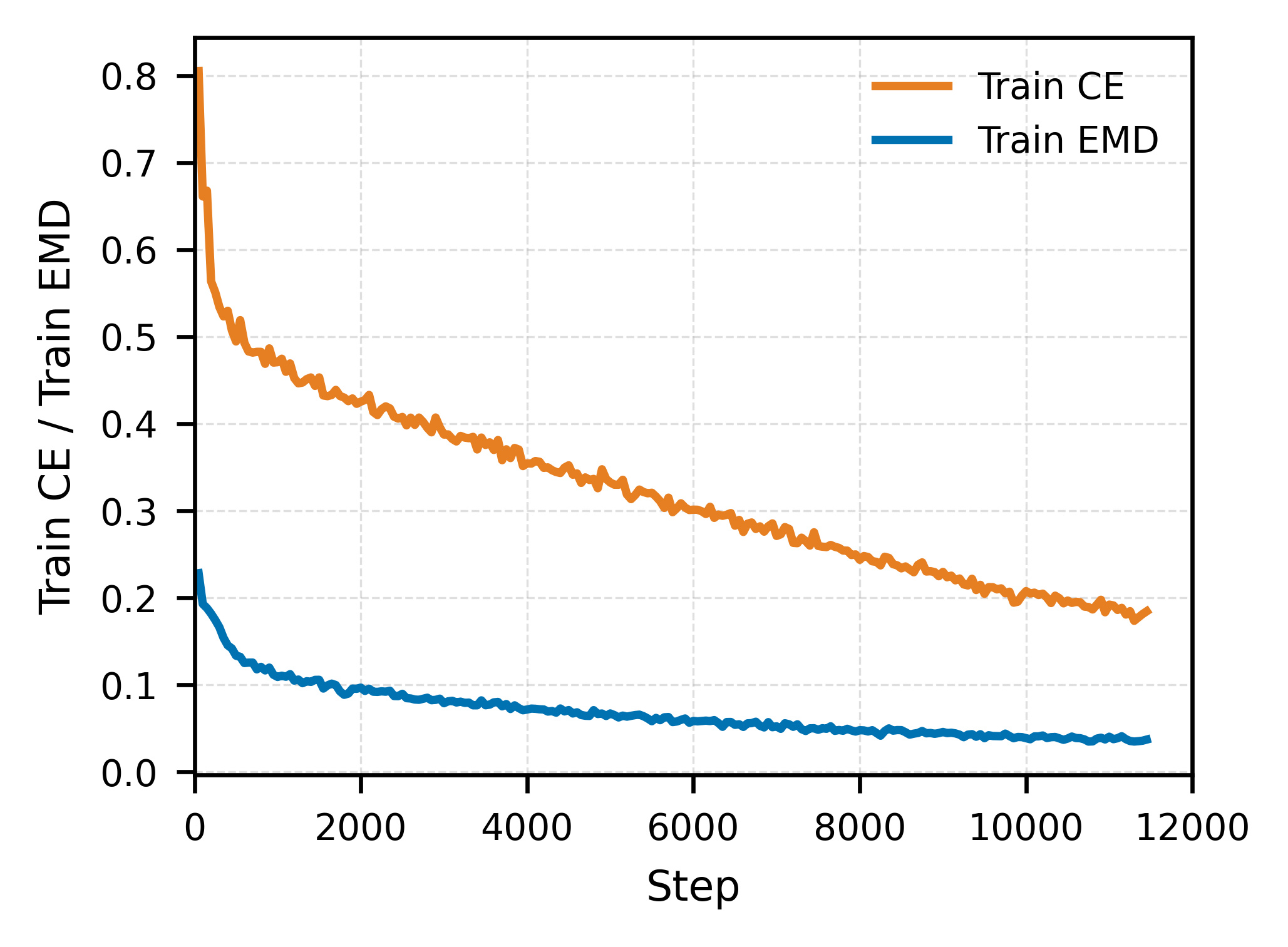}
        \caption{Training CE and EMD loss curves.}
        \label{fig:ce_vs_emd}
    \end{subfigure}
    \hfill
    \begin{subfigure}[t]{0.48\textwidth}
        \centering
        \includegraphics[width=\textwidth]{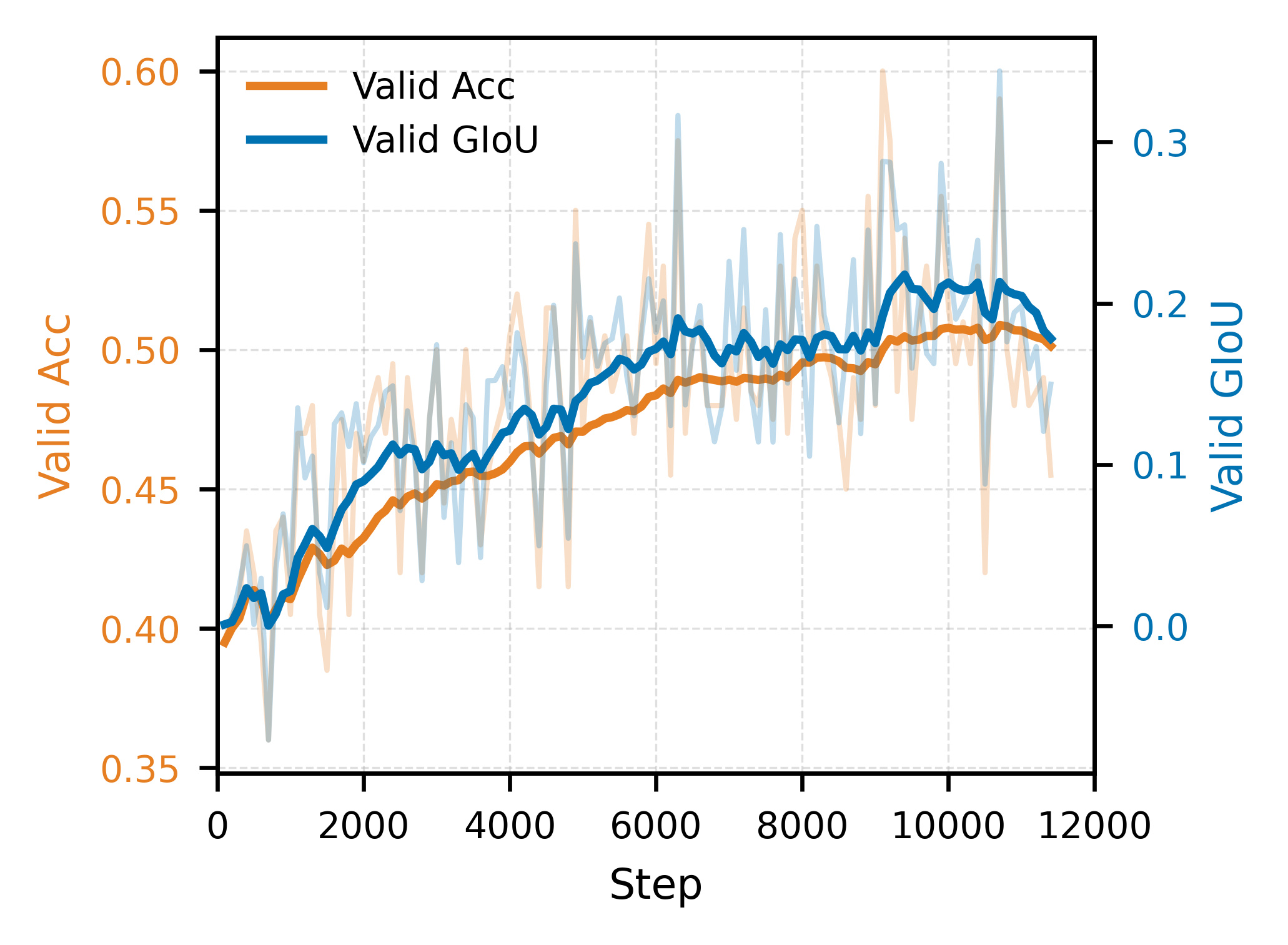}
        \caption{Validation accuracy and GIoU during training.}
        \label{fig:acc_vs_giou}
    \end{subfigure}
    \caption{\textbf{Training dynamics of our model.}
    (a) Training losses (CE and EMD) consistently decrease and stabilize, showing smooth optimization behavior.
    (b) Validation Acc and GIoU are optimized during training and eventually converge as the number of steps increases.}
    \label{fig:training_dynamics}
\end{figure*}

\cref{fig:training_dynamics}(a) illustrates the training Cross-Entropy (CE) loss and Earth Mover's Distance (EMD) loss. Both losses decrease steadily without noticeable instability or divergence, confirming that the combined objective provides stable gradients and that the optimization process converges reliably. \cref{fig:training_dynamics}(b) shows the evolution of validation accuracy and validation GIoU over training steps. Both metrics improve at the beginning of training and then gradually saturate, indicating that the model learns effective localization behavior and subsequently converges. In practice, our model converges after roughly 5 scene-centric epochs ($\approx$12 text-centric, $\sim$10{,}000 steps), still fewer than fully-supervised baselines such as 3D-VisTA~\cite{zhu20233d_vista} ($\sim$100 epochs).

\subsection{Test Time Augmentation}
\label{sec:tta}

\begin{figure}[!t]
    \centering
    \includegraphics[width=0.5\columnwidth]{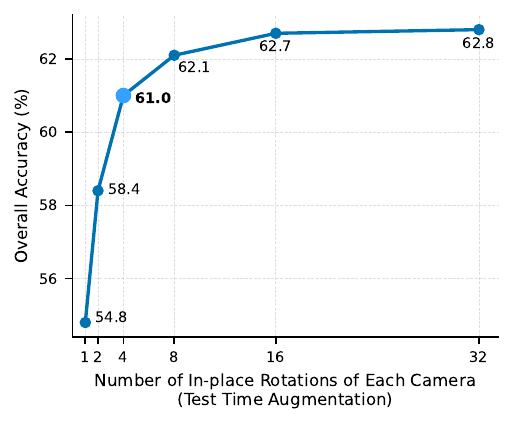}
    \caption{\textbf{Effect of test-time in-place rotations.} We evaluate how many in-place rotations per camera are used for test-time augmentation. Performance increases rapidly from 1 to 4 rotations, and gradually saturates beyond 8, indicating that \textbf{four} rotations provide the best balance between performance and computational cost.}
    \label{fig:test_time_aug}
\end{figure}

Our inference pipeline employs test-time augmentation by rendering multiple panoramas from each camera location with fixed yaw intervals. In this section, we analyze the efficacy of this strategy and justify our design choice of performing four $90^\circ$ rotations.

\cref{fig:test_time_aug} illustrates the impact of increasing the number of yaw rotations on accuracy. Performance improves significantly with fewer rotations and continues to rise until approximately 4--8 rotations, beyond which the gains saturate. This trend indicates that \textbf{four} in-place rotations offer the optimal trade-off between accuracy improvement and computational cost.

This approach can be interpreted through the lens of \emph{self-consistency}~\cite{wang2023selfconsistency}, where agreement across multiple perturbed inputs serves as a proxy for confidence. By feeding the VLM with multiple panoramas rendered from different yaw angles at the same location, we assess the reliability of the predictions based on the consistency of the model's outputs.

\subsection{Per-Category Analysis}

\begin{figure}[!t]
\begin{minipage}[t]{0.48\columnwidth}
    \centering
    \includegraphics[width=\linewidth]{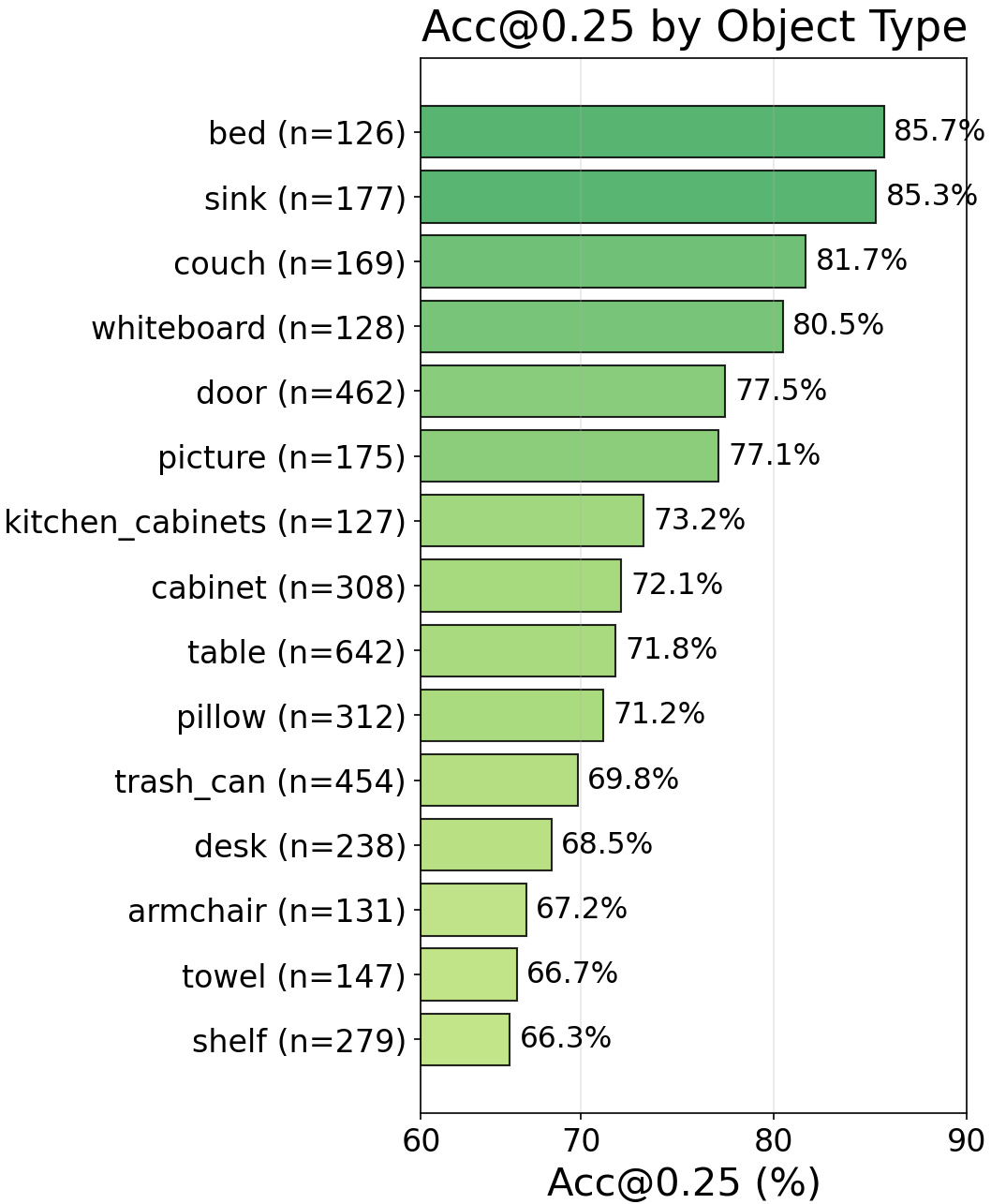}
    \captionof{figure}{\textbf{Acc@0.25 by object category}. We report Acc@0.25 for frequently mentioned categories in ScanRefer~\cite{chen2020scanrefer}. Large, distinctive objects (e.g., bed, couch) achieve the highest accuracy.}
    \label{fig:acc_by_object_type}
\end{minipage}
\hfill
\begin{minipage}[t]{0.48\columnwidth}
    \centering
    \includegraphics[width=\linewidth]{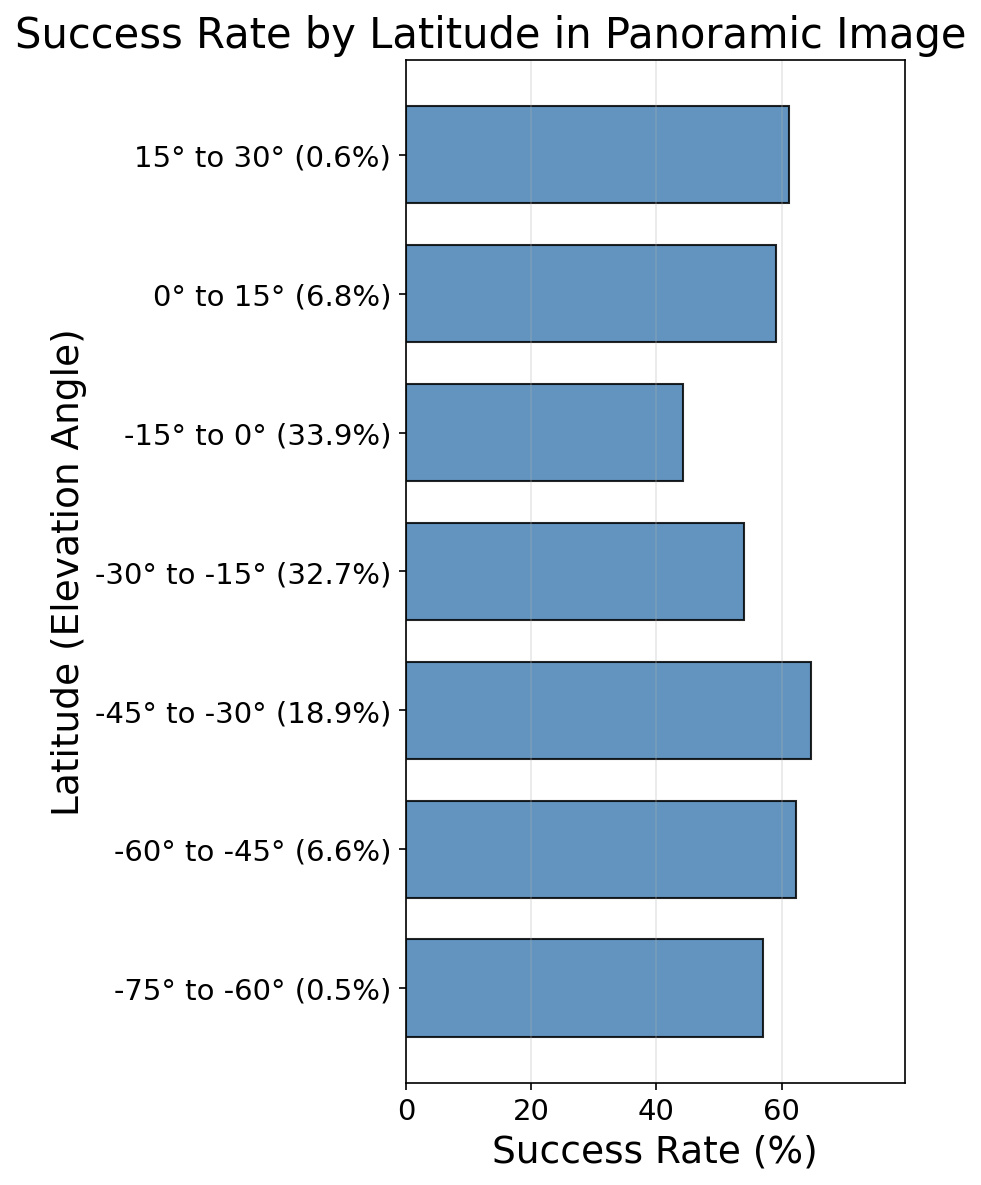}
    \captionof{figure}{\textbf{Success rate by latitude (elevation angle) in the panoramic image.} The majority of objects (66.6\%) lie between $-45^\circ$ and $0^\circ$, and accuracy is highest in the $-60^\circ$ to $-30^\circ$ range.}
    \label{fig:success_by_latitude}
\end{minipage}
\end{figure}

\cref{fig:acc_by_object_type} breaks down Acc@0.25 by 15 frequently mentioned object categories in ScanRefer. Large and visually distinctive objects such as \emph{bed} (85.7\%) and \emph{couch} (81.7\%) achieve the highest accuracy, as they are easily identifiable from panoramic views. Even mid-range categories like \emph{table} (71.8\%), \emph{trash can} (69.8\%), and \emph{desk} (68.5\%) maintain accuracy well above 65\%, indicating that our method performs robustly across a wide range of common indoor object types.

\subsection{Success Rate by Elevation Angle}

\cref{fig:success_by_latitude} analyzes grounding accuracy as a function of the object's elevation angle in the equirectangular panorama. The $-60^\circ$ to $-30^\circ$ range achieves the highest success rates, as objects in this band are at an ideal distance from the camera---close enough to be well-resolved yet not so close as to be severely distorted. In contrast, objects near the equator ($-15^\circ$ to $0^\circ$), despite being the most numerous, show slightly lower accuracy; at near-horizontal elevations the panorama captures distant regions of the scene, causing target objects to appear small and harder to localize. At very low elevations ($-75^\circ$ to $-60^\circ$), equirectangular distortion is more severe, yet accuracy remains comparable because our camera placement strategy prevents cameras from being placed directly above objects, and fine-tuning on panoramic data helps the model tolerate the remaining distortion.

\section{Efficiency and Deployment}

\subsection{Backbone Scalability and Runtime}

\begin{table}[t]
  \centering
  \caption{\textbf{Backbone scalability and resource usage on ScanRefer.} We report Acc@0.25, peak GPU memory (GB), and per-query online latency (2D inference plus 3D aggregation, in s) for three VLM backbones (Qwen2.5-VL-3B/7B and CogVLM-17B) and the LLaVA-3D~\cite{zhu2025llava3d} baseline.}
  \label{tab:backbone}
  \small
  \begin{tabular}{l l c c c}
  \toprule
  Method & Backbone & Overall & GPU Mem. & Time/Query \\
  \midrule
  LLaVA-3D & Vicuna-7B & 50.1 & 16.2 & 1.21 \\
  Ours & Qwen2.5-VL-3B & 45.6 & \first{7.0} & \first{0.72} \\
  Ours & Qwen2.5-VL-7B & 52.1 & 15.5 & 0.82 \\
  Ours & CogVLM-17B & \first{61.0} & 51.6 & 3.30 \\
  \bottomrule
  \end{tabular}
\end{table}

We study how PanoGrounder scales with model capacity by running the full pipeline on the ScanRefer~\cite{chen2020scanrefer} validation split with three VLM backbones (Qwen2.5-VL-3B/7B and CogVLM-17B). For each scene, our placement selects on average 2.4 cameras, and we render four yaw-rotated panoramas per camera, yielding $2.4 \times 4 = 9.6$ forward passes per query. We measure latency over the online components only (VLM inference and visibility-aware 3D aggregation), as camera placement and rendering are offline. All experiments run on a single NVIDIA A100 GPU. For LLaVA-3D, we measure latency and memory under the same setup and take its accuracy from the original paper.

\cref{tab:backbone} reports the resulting accuracy and per-query cost, and compares against LLaVA-3D~\cite{zhu2025llava3d} at 7B parity. Our Qwen2.5-VL-7B variant outperforms LLaVA-3D on all three axes (52.1 vs.\ 50.1 Acc@0.25, 15.5 vs.\ 16.2\,GB, 0.82 vs.\ 1.21\,s), and even surpasses dedicated 3D models such as 3D-VisTA~\cite{zhu20233d_vista} (45.9\%) and ViL3DRel~\cite{chen2022vil3drel} (47.9\%). Scaling down to Qwen2.5-VL-3B further cuts latency and memory (0.72\,s, 7.0\,GB) at some accuracy cost (45.6), while our default CogVLM-17B reaches 61.0 at higher latency (3.30\,s). Together, these results indicate that PanoGrounder's gains stem largely from its panoramic multi-modal design rather than backbone capacity alone.

\subsection{Real-World Deployment on a Self-Captured Scene}

Our benchmark experiments all assume a clean, pre-built 3D reconstruction. To test whether PanoGrounder also runs on raw user-captured data, we deployed the full pipeline end-to-end on a $4.3\times9.1$\,m living-room scene recorded with a 4\,min iPhone~11 video ($1920\times1080$, 30\,fps, RGB only), using no manual cleanup and no ground-truth camera trajectories. From the raw video, we recover camera poses and a sparse point cloud with COLMAP~\cite{schoenberger2016sfm, schoenberger2016mvs}, train a 3DGS~\cite{kerbl20233dgs} model for 30\,K steps, and then apply our standard offline pipeline (Sec.~3). Per-view 2D inference uses the checkpoint trained on 3DGS-rendered panoramas.

\begin{figure}[t]
    \centering
    \includegraphics[width=0.49\linewidth]{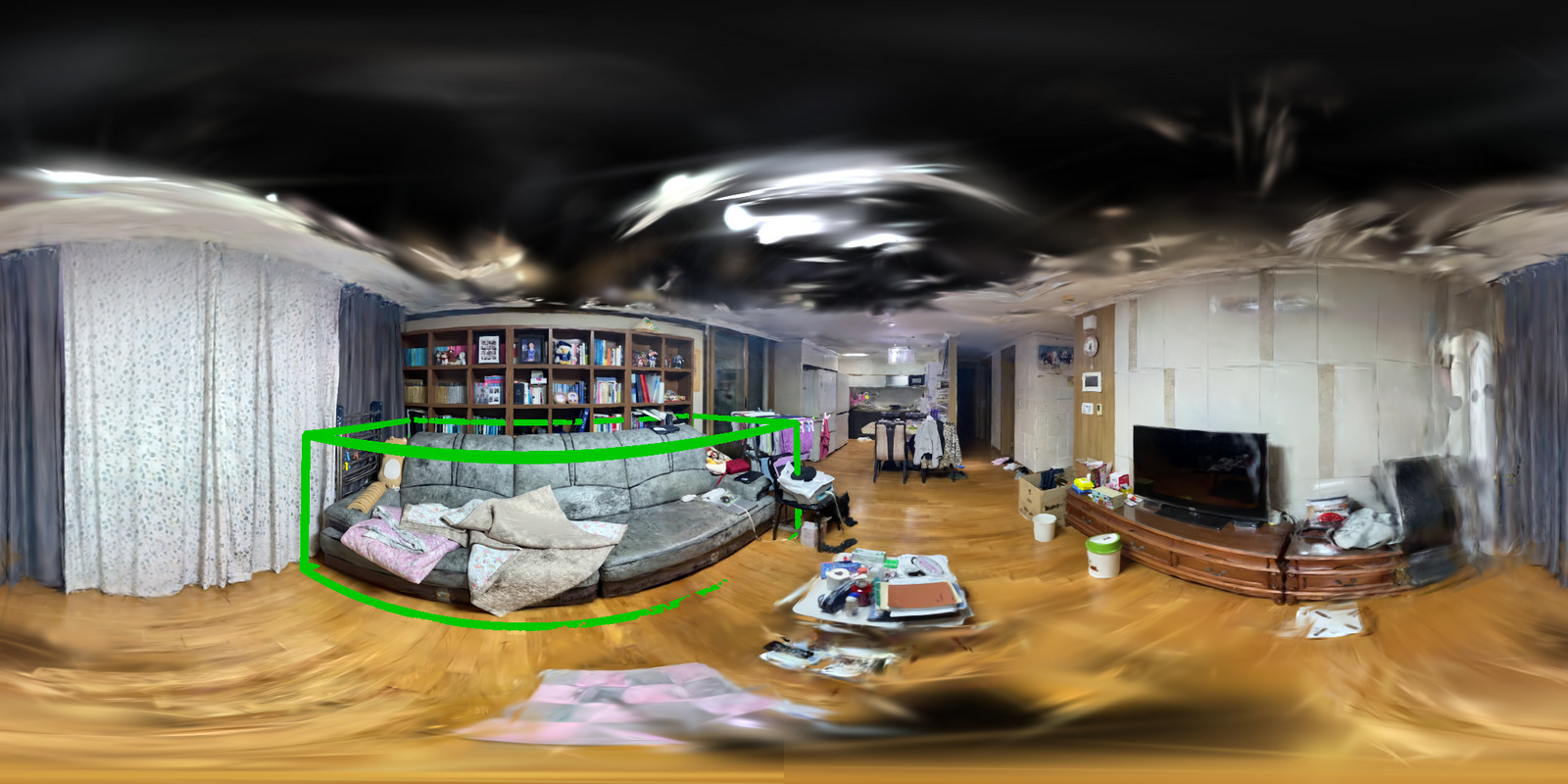}\hfill
    \includegraphics[width=0.49\linewidth]{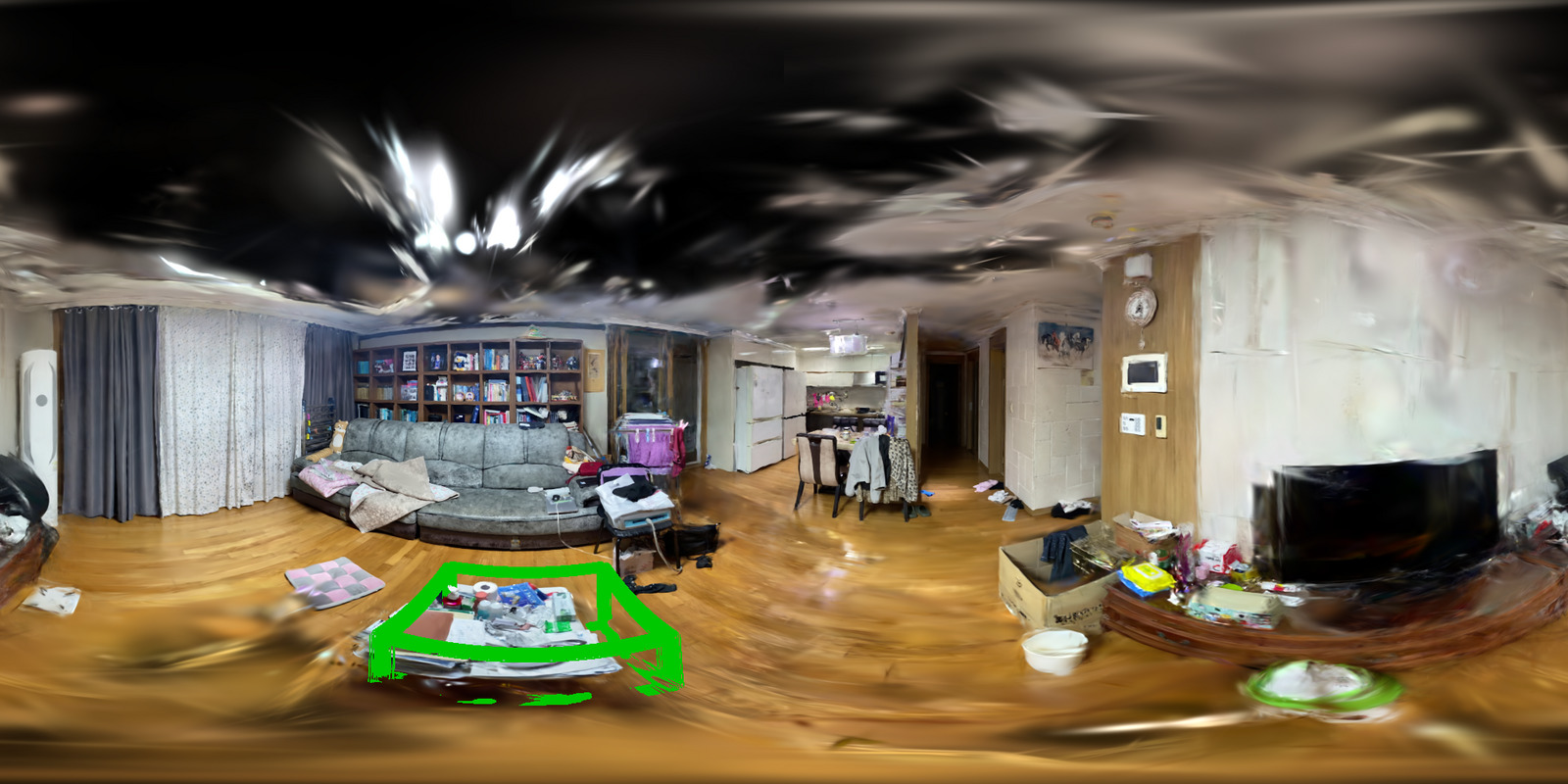}
    \caption{\textbf{End-to-end grounding on a self-captured scene.} Rendered panoramas overlaid with the predicted 3D box (green). \textbf{Left:} \emph{``\ldots gray sofa with a blanket placed on top\ldots behind the sofa, a bookshelf; to the right, a laundry drying rack.''} \textbf{Right:} \emph{``\ldots square table with a white top and wooden legs\ldots many small objects cluttered on top.''}}
    \label{fig:iphone_deploy}
\end{figure}

\cref{fig:iphone_deploy} shows two results. Both queries are long, relational expressions that mix appearance cues (color, material, shape) with spatial relations to nearby objects (``behind'', ``to the right'', ``on top''). PanoGrounder localizes the referent correctly in both, despite the imperfect reconstruction and rendering artifacts. \cref{tab:deploy_cost} reports the per-stage time and memory, grouped into one-time offline preprocessing and per-query online inference. The offline wall-clock is dominated by off-the-shelf 3D reconstruction tools---COLMAP (87\,min) and 3DGS training (43\,min). PanoGrounder's own camera placement and rendering take under two minutes. Online inference per query comprises a VLM forward pass (3.09\,s) and visibility-aware 3D aggregation (0.27\,s). Overall, these results support the practical deployability of PanoGrounder: it runs on raw, casually captured scenes with only a one-time offline setup and a few seconds of online inference per query.

\begin{table}[t]
\centering
\caption{\textbf{End-to-end cost on the self-captured household scene}, separating offline preprocessing (performed once per scene and reused across all queries) from online per-query inference.}
\label{tab:deploy_cost}
\small
\setlength{\tabcolsep}{6pt}
\begin{tabular}{lrrr}
\toprule
Stage & Time & CPU (GB) & GPU (GB) \\
\midrule
\multicolumn{4}{l}{\textbf{Offline} \emph{(once per scene; reused across all queries)}} \\
COLMAP (exhaustive matcher)  & 87\,min  & 67.1 &  2.3 \\
3DGS (30K steps)             & 43\,min  & 77.5 & 28.5 \\
Sem.\ feat.\ lift (Sec.~3.2) & 24\,min  & 80.7 & 13.7 \\
Cam.\ placement (Sec.~3.1)   & 16\,s    & 48.6 &  0.5 \\
Pano.\ render (Sec.~3.2)     & 51\,s    & 65.5 & 15.6 \\
\midrule
\multicolumn{4}{l}{\textbf{Online} \emph{(per query)}} \\
VLM forward (Sec.~3.2)       & 3.09\,s  & 21.1 & 52.4 \\
3D Aggregation (Sec.~3.3)    & 0.27\,s  & 12.4 &  7.4 \\
\bottomrule
\end{tabular}
\end{table}

\section{Qualitative Results}

\begin{figure*}[p]
    \centering
    \includegraphics[height=0.9\textwidth,angle=-90,origin=c]{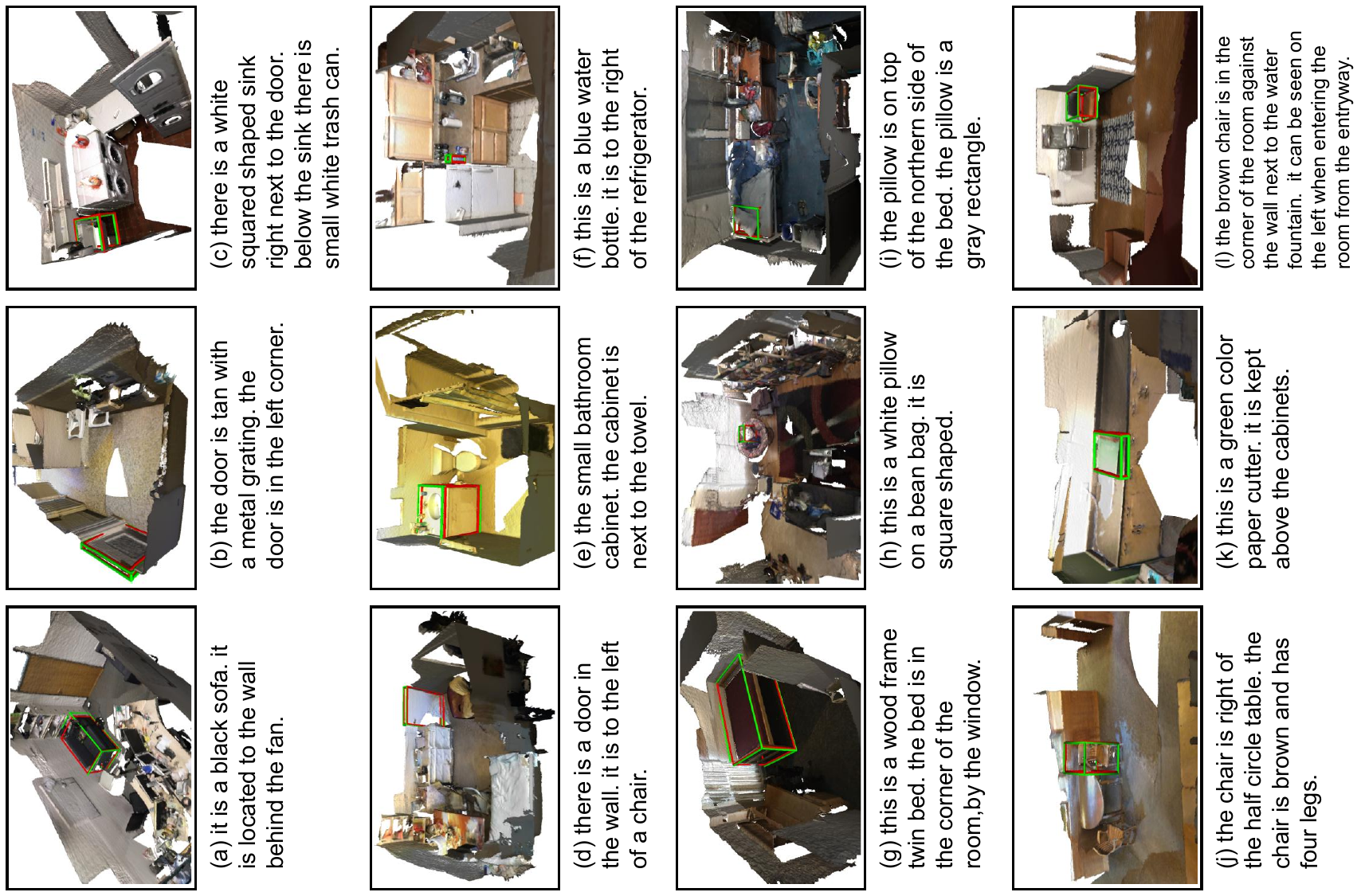}
    \caption{\textbf{Qualitative comparison with ground-truth bounding boxes.} In these examples, PanoGrounder (red boxes) produces the correct localization, similar to ground-truth boxes (green boxes).}
    \label{fig:qual_gt}
\end{figure*}

Qualitative comparisons are conducted on the ScanRefer dataset~\cite{chen2020scanrefer}. \cref{fig:qual_gt} shows representative success cases where PanoGrounder closely matches the ground-truth bounding boxes. Across a variety of room types (kitchens, bathrooms, offices, and bedrooms), the predicted boxes (red) align well with the ground-truth boxes (green), even when the target object is small (e.g., \cref{fig:qual_gt}(f), (h)). Many of the queries require understanding relational cues such as ``A next to B'' or ``A above B''. These examples illustrate that PanoGrounder can reliably parse such contextual descriptions and ground them to the correct instance in 3D.

\begin{figure*}[p]
    \centering
    \includegraphics[height=0.9\textwidth,angle=-90,origin=c]{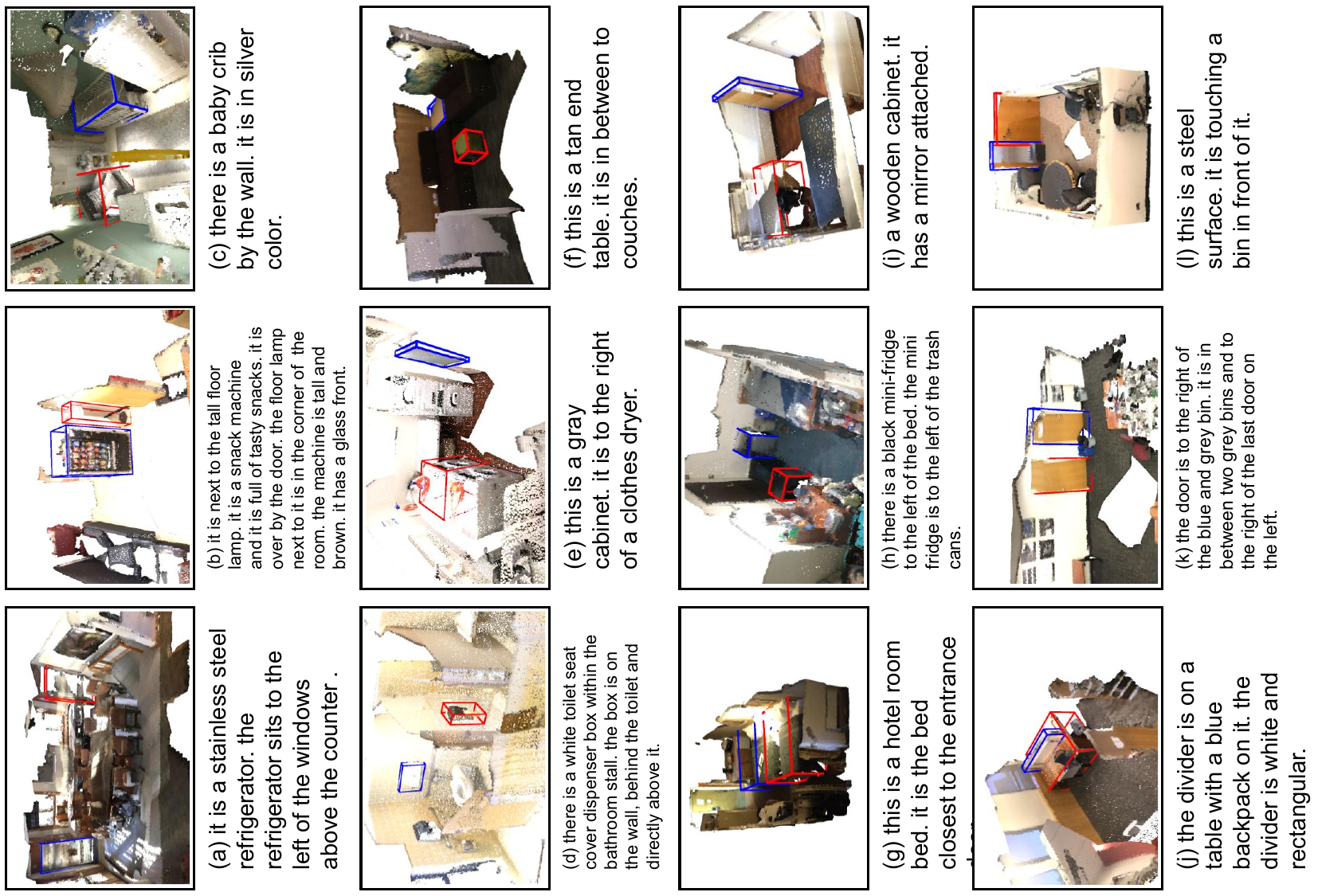}
    \caption{\textbf{Qualitative comparison with 3D-VisTA.} Both methods are provided with the same ground-truth instance masks for fair input conditions. In these examples, PanoGrounder (blue boxes) produces the correct localization, while 3D-VisTA (red boxes) yields an incorrect bounding box.}
    \label{fig:qual_3dvista}
\end{figure*}

\cref{fig:qual_3dvista} compares PanoGrounder with 3D-VisTA~\cite{zhu20233d_vista} under identical input conditions, where both methods receive the same ground-truth instance masks. 3D-VisTA (red boxes) often locks onto a plausible but incorrect instance or drifts to a nearby distractor. In contrast, thanks to the pretrained knowledge of the underlying VLM, PanoGrounder is robust on rare objects, such as those in \cref{fig:qual_3dvista}(c), (j). These qualitative results echo our quantitative findings, showing that combining panoramic context with pretrained VLM leads to more robust and precise 3D visual grounding.

\subsection{Cross-Dataset Qualitative Results}

\cref{fig:arkit_success} presents qualitative success cases on ARKitScenes~\cite{dehghan2021arkitscenes}, a dataset unseen during training, paired with human-written referring expressions from SceneVerse~\cite{jia2024sceneverse}. Despite the domain gap between the ScanNet training scenes and ARKitScenes environments, PanoGrounder successfully localizes the referred objects across diverse room types. The queries require complex spatial reasoning, such as chained relative positioning in \cref{fig:arkit_success}(a), spatial relations with room landmarks in \cref{fig:arkit_success}(c,\,d), and contextual descriptions involving nearby objects in \cref{fig:arkit_success}(e,\,f). Note that ARKitScenes does not provide axis-aligned scene coordinates, which can cause the projected 3D bounding boxes to appear tilted in the rendered views. These results complement the quantitative scene generalization findings in Sec.~4.3 of the main paper, demonstrating that our panoramic representation and VLM-based reasoning transfer effectively to novel indoor environments.

\subsection{Failure Case Analysis}

\begin{figure*}[p]
    \centering
    \includegraphics[width=0.8\textwidth,keepaspectratio]{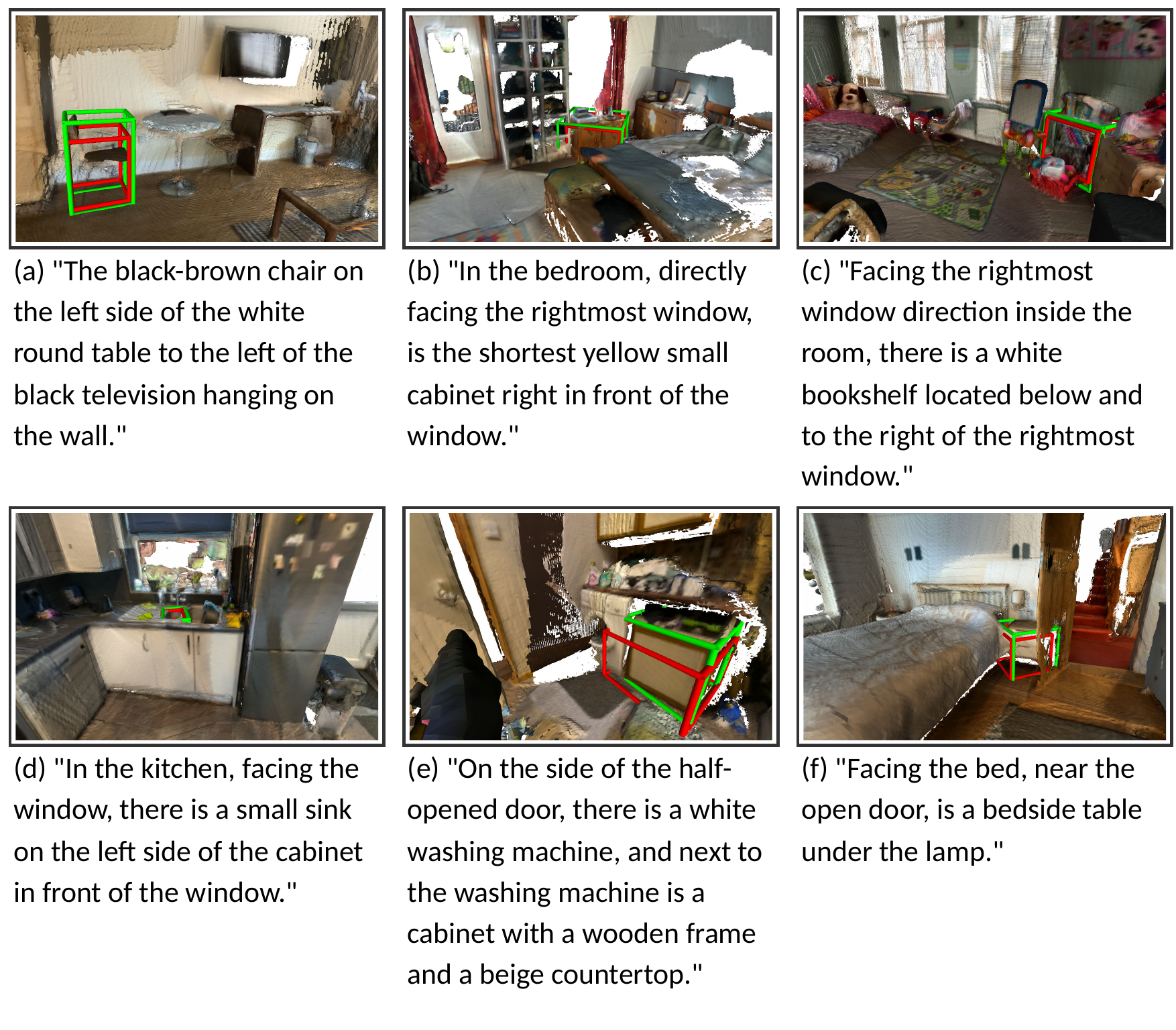}
    \caption{\textbf{Qualitative success cases on ARKitScenes~\cite{dehghan2021arkitscenes}.} Green boxes denote ground-truth and red boxes denote PanoGrounder predictions. Despite the unseen domain, PanoGrounder correctly localizes the referred objects across diverse room types.}
    \label{fig:arkit_success}

    \medskip
    \includegraphics[width=0.8\textwidth,keepaspectratio]{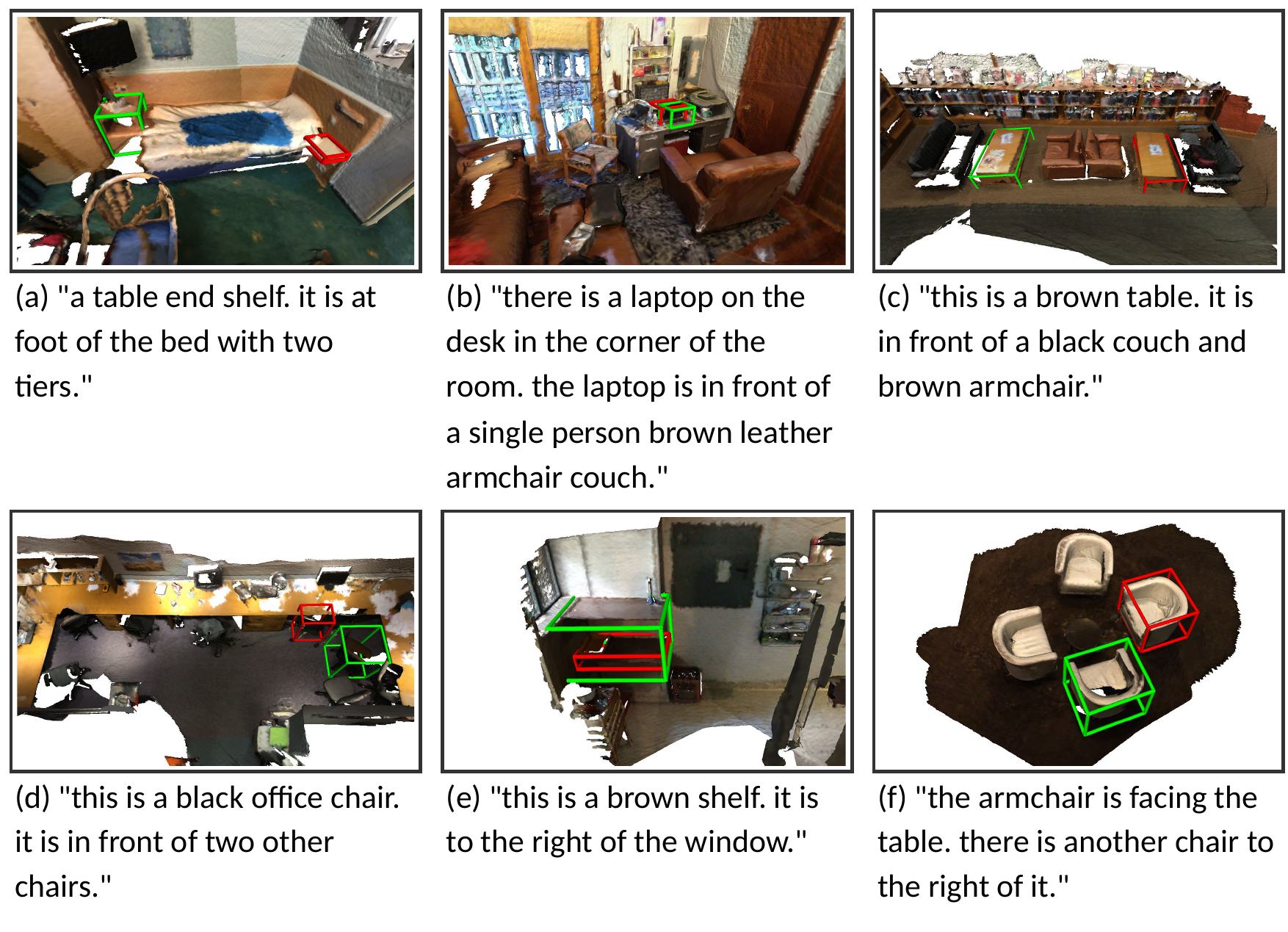}
    \caption{\textbf{Failure cases on ScanRefer~\cite{chen2020scanrefer}.} Green boxes denote ground-truth and red boxes denote PanoGrounder predictions. Common failure modes include ambiguous references with multiple plausible candidates~(c,\,d), small or occluded targets~(b), and partial localization of multi-part objects~(e).}
    \label{fig:scanrefer_failure}
\end{figure*}

\cref{fig:scanrefer_failure} illustrates representative failure cases on ScanRefer~\cite{chen2020scanrefer}. Common failure modes include: (i)~ambiguous references where multiple objects satisfy the description equally well, \eg, \cref{fig:scanrefer_failure}(c,\,d); (ii)~small or heavily occluded targets that are difficult to precisely locate in the cluttered scene, \eg, the laptop in \cref{fig:scanrefer_failure}(b); and (iii)~partial localization of multi-part objects, \eg, \cref{fig:scanrefer_failure}(e) where only the lower tier of a two-tier shelf is captured. These failure cases suggest that further improvements could be achieved by incorporating finer-grained spatial reasoning or multi-round disambiguation strategies.

\subsection{Effect of Features}

\begin{figure*}[!t]
    \centering
    \includegraphics[width=\textwidth]{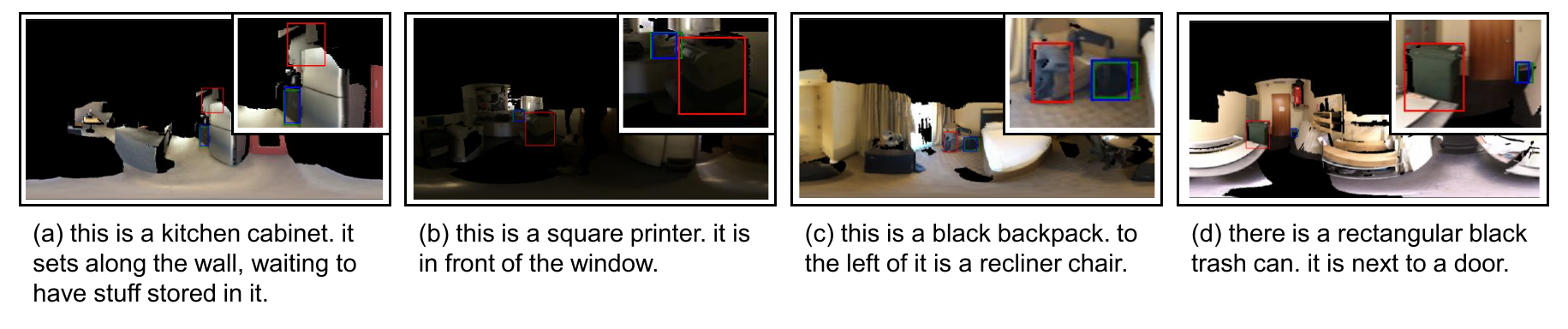}
    \caption{\textbf{Raw 2D predictions from PanoGrounder with and without semantic features.} Green boxes show ground truth, blue boxes show predictions with semantic features, and red boxes show predictions without features.}
    \label{fig:feat_abl}
\end{figure*}

In \cref{fig:feat_abl}, we compare raw 2D predictions from PanoGrounder when it is run on the RGB alone (red) versus when we augment it with our semantic features (blue). In all four examples, the blue boxes align closely with the ground-truth boxes (green), while the red boxes often drift to a nearby but incorrect region. This shows that the semantic features provide strong additional cues beyond the RGB appearance of the equirectangular image.

In \cref{fig:feat_abl}(a), the kitchen cabinet is seen from an oblique angle and is partially occluded, making it difficult to distinguish from nearby structures using appearance alone; with semantic features, PanoGrounder localizes the correct cabinet region.
In \cref{fig:feat_abl}(b), the scene is extremely dark and the printer is barely visible, so the RGB-only prediction snaps to a nearby larger object that is easier to see, whereas the semantic features guide the model toward the true printer location.
Finally, in \cref{fig:feat_abl}(d), the referred trash can is small and far from the camera; the baseline is attracted to a nearby large distractor, while the semantic features guide the prediction to the correct instance next to the door. Overall, these cases illustrate that semantic features make PanoGrounder substantially more robust to occlusion, illumination changes, textural clutter, and tiny targets.

\end{document}